\pdfoutput=1
\documentclass[letterpaper, 10 pt, conference]{ieeeconf}  

\IEEEoverridecommandlockouts                              
\makeatletter
\let\NAT@parse\undefined
\makeatother

\usepackage[numbers,sort]{natbib}
\usepackage{times}
\usepackage{epsfig}
\usepackage{graphicx}
\usepackage{amsmath}
\usepackage{amssymb}
\let\labelindent\relax 
\usepackage{enumitem}
\usepackage{color}
\usepackage{tabularx}
\usepackage{hyperref}

\usepackage[font={small}]{caption}

\newcommand{\fig}[1]{Fig.~\ref{fig:#1}}
\newcommand{\tabl}[1]{Table~\ref{table:#1}}
\newcommand{\sect}[1]{Sec.~\ref{sec:#1}}

\newcommand{\authorspace}{\qquad}

\newcommand{\faceblur}[1]{#1}

\graphicspath{{figs/}}
\usepackage{etoolbox}
\newbool{isArxiv}
\booltrue{isArxiv}

\ifbool{isArxiv}{
    \newcommand{\appen}[1]{Appendix~\ref{appendix.#1}}
    \newcommand{\afig}[1]{Fig.~\ref{fig:appendix.#1}}
    
    \newcommand{\atabl}[1]{Table~\ref{table:appendix.#1}}
    \newcommand{\acite}[1]{\cite{#1}}

    \thispagestyle{plain}
    \pagestyle{plain}
}{ 
    \newbool{isICRAtemp}
    \boolfalse{isICRAtemp}
    \newcommand{\appen}[1]{Appendix~\ref*{appendix.#1} of \citep{TCN2017}}
    \newcommand{\afig}[1]{Fig.~\ref*{fig:appendix.#1} of \citep{TCN2017}}
    
    \newcommand{\atabl}[1]{Table~\ref*{table:appendix.#1} of \citep{TCN2017}}
    \newcommand{\acite}[1]{} 

    \AtBeginEnvironment{thebibliography}{\linespread{0.75}\selectfont}
    \linespread{.91}\selectfont
}

\title{\LARGE \bf
Time-Contrastive Networks: Self-Supervised Learning from Video}
\DeclareRobustCommand*{\IEEEauthorrefmark}[1]{%
  \raisebox{0pt}[0pt][0pt]{\textsuperscript{\footnotesize #1}}%
}
\author{
Pierre Sermanet\IEEEauthorrefmark{1}\IEEEauthorrefmark{*}\IEEEauthorrefmark{$@$}\thanks{\IEEEauthorrefmark{*} equal contribution}\thanks{\IEEEauthorrefmark{$@$} correspondence to sermanet@google.com}\authorspace Corey Lynch\IEEEauthorrefmark{1}\IEEEauthorrefmark{R}\IEEEauthorrefmark{*}\thanks{\IEEEauthorrefmark{R} Google Brain Residency program (\href{http://g.co/brainresidency}{g.co/brainresidency})}\authorspace Yevgen Chebotar\IEEEauthorrefmark{2}\IEEEauthorrefmark{*}
\\
Jasmine Hsu\IEEEauthorrefmark{1}\authorspace Eric Jang\IEEEauthorrefmark{1}\authorspace Stefan Schaal\IEEEauthorrefmark{2}
\authorspace Sergey Levine\IEEEauthorrefmark{1}
\\[0.3em]
\IEEEauthorrefmark{1}Google Brain \authorspace
\IEEEauthorrefmark{2}University of Southern California
}

\begin{document}

\setlength{\abovedisplayskip}{3pt}
\setlength{\belowdisplayskip}{3pt}

\maketitle

\begin{abstract}
We propose a self-supervised approach for learning representations and robotic behaviors entirely from unlabeled videos recorded from multiple viewpoints, and study how this representation can be used in two robotic imitation settings: imitating object interactions from videos of humans, and imitating human poses. Imitation of human behavior requires a viewpoint-invariant representation that captures the relationships between end-effectors (hands or robot grippers) and the environment, object attributes, and body pose.
We train our representations using a metric learning loss, where multiple simultaneous viewpoints of the same observation are attracted in the embedding space, while being repelled from temporal neighbors which are often visually similar but functionally different. In other words, the model simultaneously learns to recognize what is common between different-looking images, and what is different between similar-looking images. This signal causes our model to discover attributes that do not change across viewpoint, but do change across time, while ignoring nuisance variables such as occlusions, motion blur, lighting and background.
We demonstrate that this representation can be used by a robot to directly mimic human poses without an explicit correspondence, and that it can be used as a reward function within a reinforcement learning algorithm. While representations are learned from an unlabeled collection of task-related videos, robot behaviors such as pouring are learned by watching a single 3rd-person demonstration by a human. Reward functions obtained by following the human demonstrations under the learned representation enable efficient reinforcement learning that is practical for real-world robotic systems. Video results, open-source code and dataset are available at \href{https://sermanet.github.io/imitate/}{sermanet.github.io/imitate}
\end{abstract}


\section{Introduction}

While supervised learning has been successful on a range of tasks where labels can be easily specified by humans, such as object classification, many problems that arise in interactive applications like robotics are exceptionally difficult to supervise. For example, it would be impractical to label every aspect of a pouring task in enough detail to allow a robot to understand all the task-relevant properties. Pouring demonstrations could vary in terms of background, containers, and viewpoint, and there could be many salient attributes in each frame, e.g. whether or not a hand is contacting a container, the tilt of the container, or the amount of liquid currently in the target vessel or its viscosity. Ideally, robots in the real world would be capable of two things: learning the relevant attributes of an object interaction task purely from observation, and understanding how human poses and object interactions can be mapped onto the robot, in order to imitate directly from \textit{third-person video observations}.

\begin{figure}[tb!]
\begin{center}
\centerline{\includegraphics[width=\columnwidth]{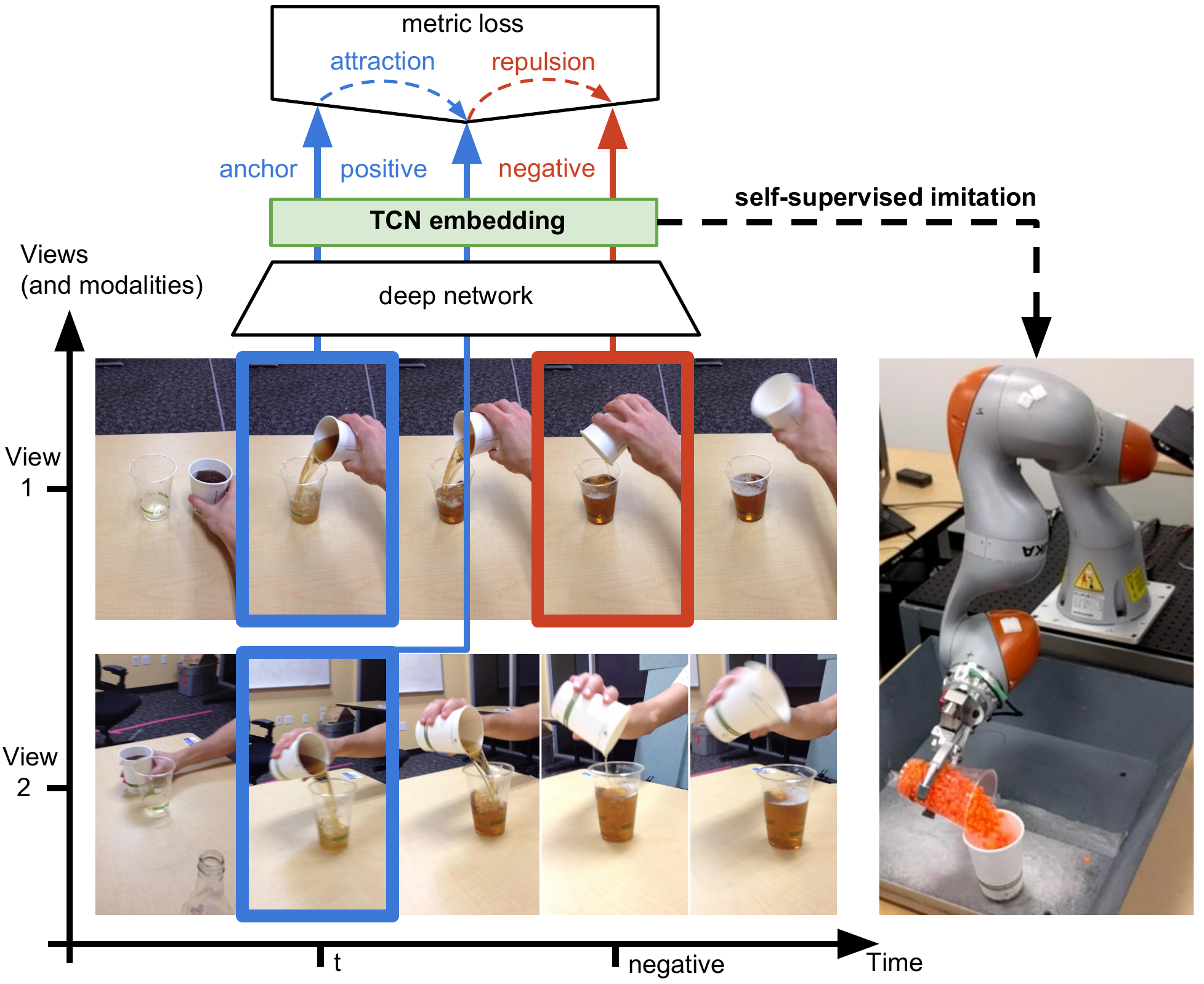}}
\caption{
{\bf Time-Contrastive Networks (TCN):} Anchor and positive images taken from simultaneous viewpoints are encouraged to be close in the embedding space, while distant from negative images taken from a different time in the same sequence. The model trains itself by trying to answer the following questions simultaneously: What is common between the different-looking blue frames? What is different between the similar-looking red and blue frames? The resulting embedding can be used for self-supervised robotics in general, but can also naturally handle 3rd-person imitation.
}
\label{fig:temporal_triplet_pouring}
\end{center}
\end{figure} 

In this work, we take a step toward addressing these challenges simultaneously through the use of self-supervision and multi-viewpoint representation learning. We obtain the learning signal from unlabeled multi-viewpoint videos of interaction scenarios, as illustrated in Figure~\ref{fig:temporal_triplet_pouring}. By learning from multi-view videos, the learned representations effectively disentangle functional attributes such as pose while being viewpoint and agent invariant. We then show how the robot can learn to link this visual representation to a corresponding motor command using either reinforcement learning or direct regression, effectively learning new tasks by observing humans.

The main contribution of our work is a representation learning algorithm that builds on top of existing semantically relevant features (in our case, features from a network trained on the ImageNet dataset~\cite{imagenet_cvpr09,DBLP:journals/corr/SzegedyVISW15}) to produce a metric embedding that is sensitive to object interactions and pose, and insensitive to nuisance variables such as viewpoint and appearance. We demonstrate that this representation can be used to create a reward function for reinforcement learning of robotic skills, using only raw video demonstrations for supervision, and for direct imitation of human poses, without any explicit joint-level correspondence and again directly from raw video. Our experiments demonstrate effective learning of a pouring task with a real robot, moving plates in and out of a dish rack in simulation, and real-time imitation of human poses. Although we train a different TCN embedding for each task in our experiments, we construct the embeddings from a variety of demonstrations in different contexts, and discuss how larger multi-task embeddings might be constructed in future work.

\section{Related Work}

\noindent {\bf Imitation learning:}
Imitation learning~\citep{ArgallChVeBr09} has been widely used for learning robotic skills from expert demonstrations~\citep{dmps, RatliffBS07,MullingKKP12,duan2017one} and can be split into two areas: behavioral cloning and inverse reinforcement learning (IRL). Behavioral cloning considers a supervised learning problem, where  examples of behaviors are provided as state-action pairs~\citep{pomerleau1991efficient,RossGB11}. IRL on the other hand uses expert demonstrations to learn a reward function that can be used to optimize an imitation policy with reinforcement learning~\citep{an04}. Both types of imitation learning typically require the expert to provide demonstrations in the same context as the learner. In robotics, this might be accomplished by means of kinesthetic demonstrations~\citep{Calinon06SMC} or teleoperation~\citep{PastorHAS09}, but both methods require considerable operator expertise. If we aim to endow robots with wide repertoires of behavioral skills, being able to acquire those skills directly from third-person videos of humans would be dramatically more scalable. Recently, a range of works have studied the problem of imitating a demonstration observed in a different context, e.g. from a different viewpoint or an agent with a different embodiment, such as a human~\citep{stadie2017third,dragan2012online,DBLP:journals/corr/SermanetXL16}.
Liu et al.~\citep{LiuGAL17} proposed to translate demonstrations between the expert and the learner contexts to learn an imitation policy by minimizing the distance to the translated demonstrations. However, Liu et al. explicitly exclude from consideration any demonstrations with domain shift, where the demonstration is performed by a human and imitated by the robot with clear visual differences (e.g., human hands vs. robot grippers). In contrast, our TCN models are trained on a diverse range of demonstrations with different embodiments, objects, and backgrounds. This allows our TCN-based method to directly mimic human demonstrations, including demonstrations where a human pours liquid into a cup, and to mimic human poses without any explicit joint-level alignment. To our knowledge, our work is the first method for imitation of raw video demonstrations that can both mimic raw videos and handle the domain shift between human and robot embodiment.

\vspace{-0.003in}
\noindent {\bf Label-free training signals:}
Label-free learning of visual representations promises to enable visual understanding from unsupervised data, and therefore has been explored extensively in recent years.
Prior work in this area has studied unsupervised learning as a way of enabling supervised learning from small labeled datasets~\citep{adversarially}, image retrieval~\citep{7410376}, and a variety of other tasks~\citep{DBLP:journals/corr/WangG15a,DBLP:journals/corr/ZhangIE16a,denoising,DBLP:conf/cvpr/GC016}.
In this paper, we focus specifically on representation learning for the purpose of model interactions between objects, humans, and their environment, which requires implicit modeling of a broad range of factors, such as functional relationships, while being invariant to nuisance variables such as viewpoint and appearance.
Our method makes use of simultaneously recorded signals from multiple viewpoints to construct an image embedding. A number of prior works have used multiple modalities and temporal or spatial coherence to extract embeddings and features.
For example, \citep{DBLP:journals/corr/OwensIMTAF15,DBLP:journals/corr/AytarVT16} used co-occurrence of sounds and visual cues in videos to learn meaningful visual features.
\citep{DBLP:journals/corr/ZhangIE16a} also propose a multi-modal approach for self-supervision by training a network for cross-channel input reconstruction.
\citep{DBLP:journals/corr/DoerschGE15,DBLP:conf/cvpr/ZagoruykoK15} use the spatial coherence in images as a self-supervision signal and \citep{DBLP:journals/corr/PathakGDDH16} use motion cues to self-supervise a segmentation task. These methods are more focused on spatial relationships, and the unsupervised signal they provide is complementary to the one explored in this work.

A number of prior works use temporal coherence~\citep{Wiskott:2002:SFA:638940.638941,goroshin,DBLP:journals/corr/FernandoBGG16,DBLP:journals/corr/MisraZH16}. Others also train for viewpoint invariance using metric learning~\citep{DBLP:conf/cvpr/GC016,DBLP:journals/corr/YiTLF16,simo2015discriminative}. The novelty of our work is to combine both aspects in opposition, as explained in \sect{tc_signal}.
\citep{DBLP:journals/corr/WangG15a} uses a triplet loss that encourages first and last frames of a tracked sequence to be closer together in the embedding, while random negative frames from other videos are far apart. Our method differs in that we use temporal neighbors as negatives to push against a positive that is anchored by a simultaneous viewpoint. This causes our method to discover meaningful dimensions such as attributes or pose, while \citep{DBLP:journals/corr/WangG15a} focuses on learning intraclass invariance. Simultaneous multi-view capture also provides exact correspondence while tracking does not, and can provide a rich set of correspondences such as occlusions, blur, lighting and viewpoint.

Other works have proposed to use prediction as a learning signal~\citep{DBLP:journals/corr/WhitneyCKT16,journals/corr/MathieuCL15}. The resulting representations are typically evaluated primarily on the realism of the predicted images, which remains a challenging open problem. A number of other prior methods have used a variety of labels and priors to learn embeddings.
\citep{DBLP:journals/corr/MoriPKLTTY15} use a labeled dataset to train a pose embedding, then find the nearest neighbors for new images from the training data for a pose retrieval task. Our method is initialized via ImageNet training, but can discover dimensions such as pose and task progress (e.g., for a pouring task) without any task-specific labels.
\citep{DBLP:journals/corr/StewartE16} explore various types of physical priors, such as the trajectories of objects falling under gravity, to learn object tracking without explicit supervision. Our method is similar in spirit, in that it uses temporal co-occurrence, which is a universal physical property, but the principle we use is general and broadly applicable and does not require task-specific input of physical rules.

\ifbool{isArxiv}{
\noindent {\bf Mirror Neurons:}
Humans and animals have been shown, experimentally, to possess viewpoint-invariant representations of objects and other agents in their environment \cite{caggiano2011view}, and the well known work on ``mirror neurons'' has demonstrated that these viewpoint invariant representations are crucial for imitation \citep{RizzolattiCraighero04}. Our multi-view capture setup in \fig{capture} is similar to the experimental setup used by \citep{caggiano2011view}, and our robot imitation setup, where a robot imitates human motion without ever receiving ground truth pose labels, examines how self-supervised pose recognition might arise in a learned system.
}

\section{Imitation with Time-Contrastive Networks}

Our approach to imitation learning is to only rely on sensory inputs from the world. We achieve this in two steps.
First, we learn abstract representations purely from passive observation. Second, we use these representations to guide robotic imitations of human behaviors and learn to perform new tasks.
We use the term imitation rather than demonstrations because our models also learn from passive observation of non-demonstration behaviors. A robot needs to have a general understanding about everything it sees in order to better recognize an active demonstration. We purposely insist on only using self-supervision to keep the approach scalable in the real world.
In this work, we explore a few ways to use time as a signal for unsupervised representation learning. We also explore different approaches to self-supervised robotic control below.

\subsection{Training Time-Contrastive Networks}
\label{sec:tc_signal}

\begin{figure}[h]
\begin{center}
\centerline{\includegraphics[width=.6\columnwidth]{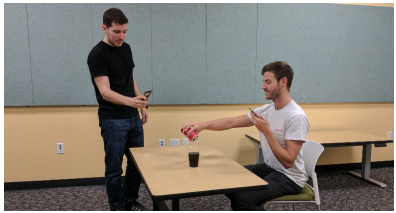}}
\caption{{\bf Multi-view capture} with two operators equipped with smartphones. Moving the cameras around freely introduces a rich variety of scale, viewpoint, occlusion, motion-blur and background correspondences between the two cameras.}
\label{fig:capture}
\end{center}
\end{figure} 

We illustrate our time-contrastive (TC) approach in \fig{temporal_triplet_pouring}. The method uses multi-view metric learning via a triplet loss~\citep{DBLP:journals/corr/SchroffKP15}. The embedding of an image $x$ is represented by $f(x)\in\mathbb{R}^d$. The loss ensures that a pair of co-occuring frames $x_i^a$ (\emph{anchor}) and $x_i^p$ (\emph{positive}) are closer to each other in embedding space than any image $x_i^n$ (\emph{negative}). Thus, we aim to learn an embedding $f$ such that
\begin{align*}
  \|f(x_i^a) - f(x_i^p)\|_2^2 + \alpha &< \|f(x_i^a) - f(x_i^n)\|_2^2\;,\\
  \:\forall\left(f(x_i^a), f(x_i^p), f(x_i^n)\right)\in\mathcal{T},
\end{align*}
where $\alpha$ is a margin that is enforced between positive and negative pairs, and $\mathcal{T}$ is the set of all possible triplets in the training set. 
The core idea is that two frames (anchor and positive) coming from the same time but different viewpoints (or modalities) are pulled together, while a visually similar frame from a temporal neighbor is pushed apart. This signal serves two purposes: learn disentangled representations without labels and simultaneously learn viewpoint invariance for imitation. The cross-view correspondence encourages learning invariance to viewpoint, scale, occlusion, motion-blur, lighting and background, since the positive and anchor frames show the same subject with variations along these factors. For example, \fig{temporal_triplet_pouring} exhibits all these transformations between the top and bottom sequences, except for occlusion.
In addition to learning a rich set of visual invariances, we are also interested in viewpoint invariance for 3rd-person to 1st-person correspondence for imitation.
How does the time-contrastive signal lead to disentangled representations? It does so by introducing competition between temporal neighbors to explain away visual changes over time. For example, in \fig{temporal_triplet_pouring}, since neighbors are visually similar, the only way to tell them apart is to model the amount of liquid present in the cup, or to model the pose of hands or objects and their interactions. Another way to understand the strong training signal that TCNs provide is to recognize the two constraints being simultaneously imposed on the model: along the view axis in \fig{temporal_triplet_pouring} the model learns to explain what is common between images that look different, while along the temporal axis it learns to explain what is different between similar-looking images. 
Note that while the natural ability for imitation of this approach is important, its capability for learning rich representations without supervision is an even more significant contribution. 
The key ingredient in our approach is that multiple views ground and disambiguate the possible explanations for changes in the physical world. We show in \sect{experiments} that the TCN can indeed discover correspondences between different objects or bodies, as well as attributes such as liquid levels in cups and pouring stages, all without supervision. This is a somewhat surprising finding as no explicit correspondence between objects or bodies is ever provided.
We hypothesize that manifolds of different but functionally similar objects naturally align in the embedding space, because they share some functionality and appearance.  

Multi-view data collection is simple and can be captured with just two operators equipped with smartphones, as shown in \fig{capture}. One operator keeps a fixed point of view of the region of interest while performing the task, while the other moves the camera freely to introduce the variations discussed above. While more cumbersome than single-view capture, we argue that multi-view capture is cheap, simple, and practical, when compared to alternatives such as human labeling.

\begin{figure}[htbp]
\begin{center}
\centerline{\includegraphics[width=\columnwidth]{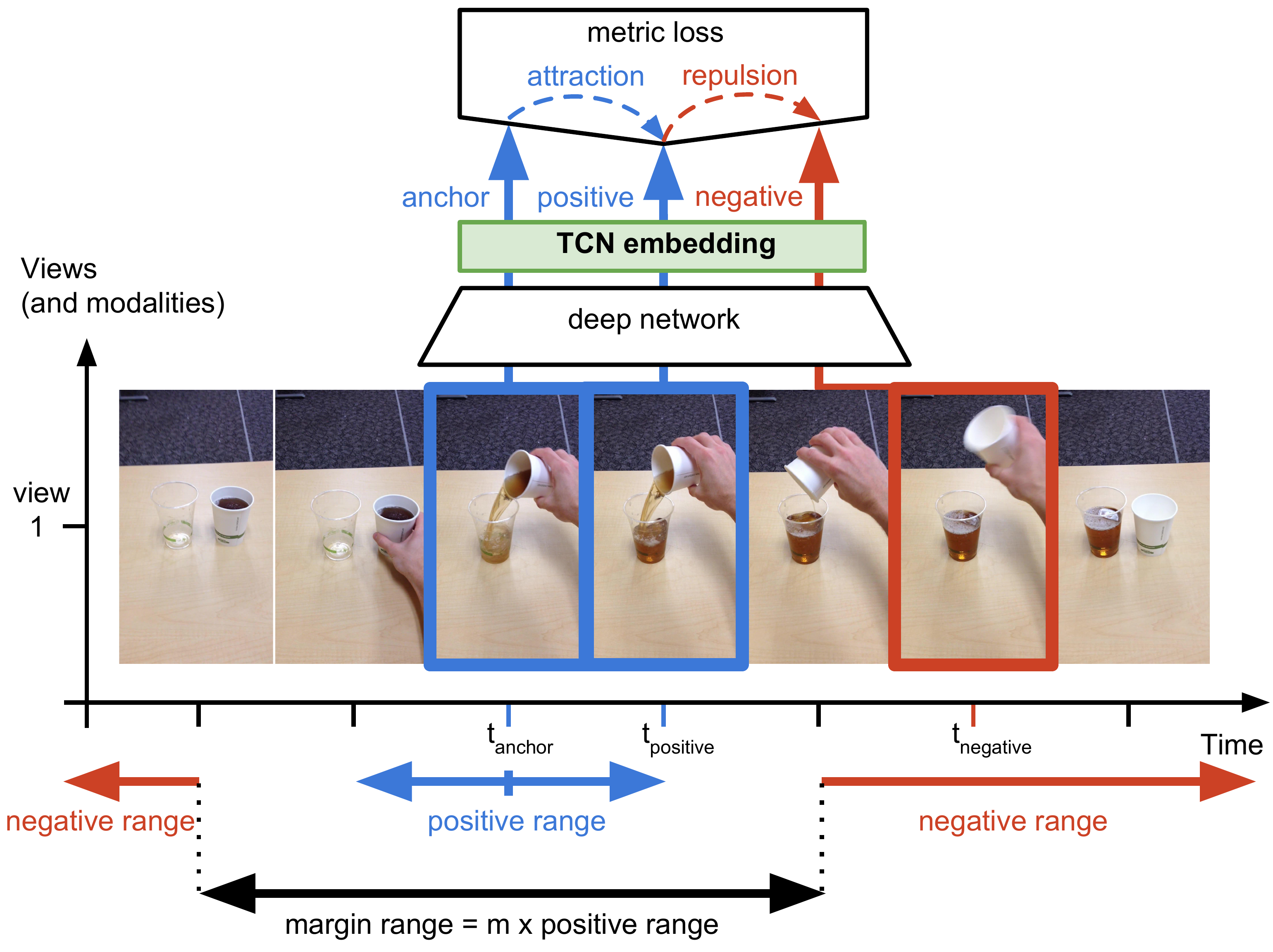}}
\caption{
{\bf Single-view TCN:} positives are selected within a small window around anchors, while negatives are selected from distant timesteps in the same sequence.}
\label{fig:sv_temporal_triplet_pouring}
\vspace{-0.5em}
\end{center}
\end{figure} 

We can also consider time-contrastive models trained on single-view video as shown in \fig{sv_temporal_triplet_pouring}. In this case, the positive frame is randomly selected within a certain range of the anchor. A margin range is then computed given the positive range. Negatives are randomly chosen outside of the margin range and the model is trained as before. We empirically chose the margin range to be $2$ times the positive range, which is itself set to $0.2s$. While we show in \sect{experiments} that multi-view TCN performs best, the single-view version can still be useful when no multi-view data is available.

\vspace{-0.05in}
\subsection{Learning Robotic Behaviors with Reinforcement Learning}
\label{sec:reward_imitation}
In this work, we consider an imitation learning scenario where the demonstrations come from a 3rd-person video observation of an agent with an embodiment that differs from the learning agent, e.g. robotic imitation of a human. Due to differences in the contexts, direct tracking of the demonstrated pixel values does not provide a sensible way of learning the imitation behavior. As described in the previous section, the TCN embedding provides a way to extract image features that are invariant to the camera angle and the manipulated objects, and can explain physical interactions in the world. We use this insight to construct a reward function that is based on the distance between the TCN embedding of a human video demonstration and camera images recorded with a robot camera. As shown in \sect{rl_experiments}, by optimizing this reward function through trial and error we are able to mimic demonstrated behaviors with a robot, utilizing only its visual input and the video demonstrations for learning. Although we use multiple multi-view videos to train the TCN, the video demonstration consists only of a \textit{single} video of a human performing the task from a random viewpoint.

Let $V=(\mathbf{v}_1,\dots \mathbf{v}_T)$ be the TCN embeddings of each frame in a video demonstration sequence. For each camera image observed during a robot task execution, we compute TCN embeddings $W=(\mathbf{w}_1,\dots \mathbf{w}_T)$. We define a reward function $R(\mathbf{v}_t, \mathbf{w}_t)$ 
based on the squared Euclidean distance and a Huber-style loss:
\begin{align*}
R(\mathbf{v}_t, \mathbf{w}_t) = -\alpha \|\mathbf{w}_t - \mathbf{v}_t\|_2^2 - \beta\sqrt{\gamma+\|\mathbf{w}_t - \mathbf{v}_t\|_2^2}
\end{align*}
where $\alpha$ and $\beta$ are weighting parameters (empirically chosen), and $\gamma$ is a small constant. The squared Euclidean distance (weighted by $\alpha$) gives us stronger gradients when the embeddings are further apart, which leads to larger policy updates at the beginning of learning. The Huber-style loss (weighted by $\beta$) starts prevailing when the embedding vectors are getting very close ensuring high precision of the task execution and fine-tuning of the motion towards the end of the training.  

In order to learn robotic imitation policies, we optimize the reward function described above using reinforcement learning. In particular, for optimizing robot trajectories, we employ the PILQR algorithm~\citep{pilqr_icml17}. This algorithm combines approximate model-based updates via LQR with fitted time-varying linear dynamics, and model-free corrections. We notice that in our tasks, the TCN embedding provides a well-behaved low-dimensional (32-dimensional in our experiments) representation of the state of the visual world in front of the robot. By including the TCN features in the system state (i.e. state = joint angles + joint velocities + TCN features), we can leverage the linear approximation of the dynamics during the model-based LQR update and significantly speed up the training. 
The details of the reinforcement learning setup can be found in \appen{rl}.

\subsection{Direct Human Pose Imitation}

In the previous section, we discussed how reinforcement learning can be used with TCNs to enable learning of object interaction skills directly from video demonstrations of humans. In this section, we describe another approach for using TCNs: direct imitation of human pose. While object interaction skills primarily require matching the functional aspects of the demonstration, direct pose imitation requires learning an implicit mapping between human and robot poses, and therefore involves a much more fine-grained association between frames. Once learned, a human-robot mapping could be used to speed up the exploration phase of RL by initializing a policy close to the solution.

\begin{figure}[h]
\begin{center}
\centerline{\includegraphics[width=\columnwidth]{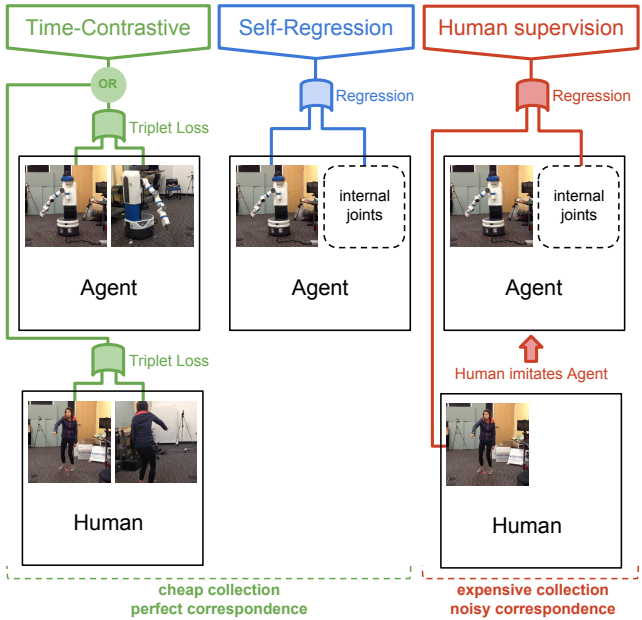}}
\caption{
{\bf Training signals for pose imitation:} time-contrastive, self-regression and human supervision.
The time-contrastive signal lets the model learn rich representations of humans or robots individually. Self-regression allows the robot to predict its own joints given an image of itself. The human supervision signal is collected from humans attempting to imitate robot poses.
}
\vspace{-1.0em}
\label{fig:signals}
\end{center}
\end{figure}

We learn a direct pose imitation through self-regression. It is illustrated in \fig{signals} and \fig{self_regression_model} in the context of self-supervised human pose imitation. The idea is to directly predict the internal state of the robot given an image of itself. Akin to looking at itself in the mirror, the robot can regress its prediction of its own image to its internal states. We first train a shared TCN embedding by observing human and robots performing random motions. Then the robot trains itself with self-regression. Because it uses a TCN embedding that is invariant between humans and robots, the robot can then naturally imitate humans after training on itself. Hence we obtain a system that can perform end-to-end imitation of human motion, even though it was never given any human pose labels nor human-to-robot correspondences. We demonstrate a way to collect human supervision for end-to-end imitation in \fig{signals}. However contrary to time-contrastive and self-regression signals, the human supervision is very noisy and expensive to collect. We use it to benchmark our approach in \sect{imitation_learning} and show that large quantities of cheap supervision can effectively be mixed with small amounts of expensive supervision.

\section{Experiments}
\label{sec:experiments}

Our experiments aim to study three questions. First, we examine whether the TCN can learn visual representations that are more indicative of object interaction attributes, such as the stages in a pouring task. This allows us to comparatively evaluate the TCN against other self-supervised representations. Second, we study how the TCN can be used in conjunction with reinforcement learning to acquire complex object manipulation skills in simulation and on a real-world robotic platform. Lastly, we demonstrate that the TCN can enable a robot to perform continuous, real-time imitation of human poses without explicitly specifying any joint-level correspondences between robots and humans. Together, these experiments illustrate the applicability of the TCN representation for modeling poses, object interactions, and the implicit correspondences between robot imitators and human demonstrators.
    
\subsection{Discovering Attributes from General Representations}
\label{sec:world_understanding}

\subsubsection{Liquid Pouring}

In this experiment, we study what the TCN captures simply by observing a human subject pouring liquids from different containers into different cups. The videos were captured using two standard smartphones (see \fig{capture}), one from a subjective point of view by the human performing the pouring, and the other from a freely moving third-person viewpoint. Capture is synchronized across the two phones using an off-the-shelf app and each sequence is approximately 5 seconds long. We divide the collected multi-view sequences into 3 sets: 133 sequences for training (about 11 minutes total), 17 for validation and 30 for testing. The training videos contain clear and opaque cups, but we restrict the testing videos to clear cups only in order to evaluate if the model has an understanding of how full the cups are.

\subsubsection{Models}
\label{sec:models}

In all subsequent experiments, we use a custom architecture derived from the Inception architecture~\citep{DBLP:journals/corr/SzegedyVISW15} that is similar to~\citep{DBLP:journals/corr/FinnTDDLA15}. It consists of the Inception model up until the layer ``Mixed\_5d'' (initialized with ImageNet pre-trained weights), followed by 2 convolutional layers, a spatial softmax layer~\citep{DBLP:journals/corr/FinnTDDLA15} and a fully-connected layer. The embedding is a fully connected layer with 32 units added on top of our custom model. This embedding is trained either with the multi-view TC loss, the single-view TC loss, or the shuffle \& learn loss~\citep{DBLP:journals/corr/MisraZH16}. For the TCN models, we use the triplet loss from~\citep{DBLP:journals/corr/SchroffKP15} without modification and with a gap value of $0.2$. Note that, in all experiments, negatives always come from the same sequence as positives. We also experiment with other metric learning losses, namely npairs \cite{sohn2016improved} and lifted structured \cite{DBLP:journals/corr/SongXJS15}, and show that results are comparable.
We use the output of the last layer before the classifier of an ImageNet-pretrained Inception model~\citep{imagenet_cvpr09,DBLP:journals/corr/SzegedyVISW15} (a 2048-dimensional vector) as a baseline in the following experiments, and call it "Inception-ImageNet".
Since the custom model is initialized from ImageNet pre-training, it is a natural point of comparison which allows us to control for any invariances that are introduced through ImageNet training rather than other approaches. 
We compare TCN models to a shuffle \& learn baseline trained on our data, using the same hyper-parameters taken from the paper (tmax of 60, tmin of 15, and negative class ratio of 0.75). Note that in our implementation, neither the shuffle \& learn baseline nor TCN benefit from a biased sampling to high-motion frames. To investigate the differences between multi-view and single-view, we compare to a single-view TCN, with a positive range of $0.2$ seconds and a negative multiplier of $2$.

\subsubsection{Model selection}

The question of model selection arises in unsupervised training. Should you select the best model based on the validation loss? Or hand label a small validation for a given task? We report numbers for both approaches. In \tabl{order_metric_selected_from_loss} we select each model based on the its lowest validation loss, while in \atabl{order_metric_selected_from_classification} we select based on a classification score from a small validation set labeled with the 5 attributes described earlier. As expected, models selected by validation classification score perform better on the classification task. However models selected by loss perform only slightly worse, except for shuffle \& learn, which suffers a bigger loss of accuracy. We conclude that it is reasonable for TCN models to be selected based on validation loss, not using any labels.

\subsubsection{Training time}

We observe in \atabl{order_metric_selected_from_classification} that the multi-view TCN (using triplet loss) outperforms single-view models while requiring 15x less training time and while being trained on the exact same dataset. We conclude that taking advantage of temporal correspondences greatly improves training time and accuracy.
\vspace{-0.2cm}
\begin{table}[h]
\small
\begin{center}
\begin{tabular}{l|c|c|c}
{\bf Method} & {\bf alignment} & {\bf classif.} & {\bf training} \\
& {\bf error} & {\bf error} & {\bf iteration}\\
\hline
Random & $28.1\% $ & $54.2\%$ & -\\
Inception-ImageNet & $ 29.8\% $ & $ 51.9\%$ & -\\
shuffle \& learn~\citep{DBLP:journals/corr/MisraZH16} & $ 22.8\% $ & $ 27.0\%$ & 575k\\
single-view TCN (triplet) & $ 25.8\% $ & $ 24.3\% $& 266k\\
multi-view TCN (npairs)& $ 18.1 \% $ & $  22.2\% $ & 938k\\
multi-view TCN (triplet)& $ 18.8\% $ & $ 21.4\% $ & 397k\\
multi-view TCN (lifted)& $18.0 \% $ & $ 19.6\% $ & 119k\\
\end{tabular}
\caption{\textbf{Pouring alignment and classification errors:} all models are selected at their lowest validation loss. The classification error considers 5 classes related to pouring detailed in \tabl{attributes_error}.}
\label{table:order_metric_selected_from_loss}
\end{center}
\end{table}

\subsubsection{Quantitative Evaluation}

We present two metrics in \tabl{order_metric_selected_from_loss} to evaluate what the models are able to capture. The alignment metric measures how well a model can semantically align two videos. The classification metric measures how well a model can disentangle pouring-related attributes, that can be useful in a real robotic pouring task.
All results in this section are evaluated using nearest neighbors in embedding space. Given each frame of a video, each model has to pick the most semantically similar frame in another video. The "Random" baseline simply returns a random frame from the second video.

The sequence alignment metric is particularly relevant and important when learning to imitate, especially from a third-party perspective. For each pouring test video, a human operator labels the key frames corresponding to the following events: the first frame with hand contact with the pouring container, the first frame where liquid is flowing, the last frame where liquid is flowing, and the last frame with hand contact with the container. These keyframes establish a coarse semantic alignment which should provide a relatively accurate piecewise-linear correspondence between all videos.
For any pair of videos $(v_1, v_2)$ in the test set, we embed each frame given the model to evaluate. For each frame of the source video $v_1$, we associate it with its nearest neighbor in embedding space taken from all frames of $v_2$. We evaluate how well the nearest neighbor in $v_2$ semantically aligns with the reference frame in $v_1$. Thanks to the labeled alignments, we find the proportional position of the reference frame with the target video $v_2$, and compute the frame distance to that position, normalized by the target segment length.

We label the following attributes in the test and validation sets to evaluate the classification task as reported in \tabl{attributes_error}: is the hand in contact with the container? (yes or no); is the container within pouring distance of the recipient? (yes or no); what is the tilt angle of the pouring container? (values 90, 45, 0 and -45 degrees); is the liquid flowing? (yes or no); does the recipient contain liquid? (yes or no). These particular attributes are evaluated because they matter for imitating and performing a pouring task. Classification results are normalized by class distribution. Note that while this could be compared to a supervised classifier, as mentioned in the introduction, it is not realistic to expect labels for every possible task in a real application, e.g. in robotics. Instead, in this work we aim to compare to realistic general off-the-shelf models that one might use without requiring new labels.

In \tabl{order_metric_selected_from_loss}, we find that the multi-view TCN model outperforms all baselines. We observe that single-view TCN and shuffle \& learn are on par for the classification metric but not for the alignment metric. We find that general off-the-shelf Inception features significantly under-perform compared to other baselines.
Qualitative examples and t-SNE visualizations of the embedding are available in \appen{objects_interaction_analysis}. We encourage readers to refer to supplementary videos to better grasp these results. 
\vspace{-0.2cm}
\begin{table}[h]
\tiny
\begin{center}
\begin{tabular}{l|c|c|c|c|c}
{\bf Method} & hand & within & container &  liquid & recipient \\
 & contact & pouring & angle & is & has \\
 & with container & distance &  & flowing & liquid \\
 & container &  &  &  & \\
\hline
Random & $49.9\%$ & $48.9\%$ & $74.5\%$ & $49.2\%$ & $48.4\%$ \\
Imagenet Inception    & $47.4\%$ & $45.2\%$ & $71.8\%$ & $48.8\%$ & $49.2\%$ \\
shuffle \& learn            & $17.2\%$ & $17.8\%$ & $46.3\%$ & $25.7\%$ & $28.0\%$ \\
single-view TCN (triplet)   & $12.6\%$ & $14.4\%$ & $41.2\%$ & $21.6\%$ & $31.9\%$ \\
multi-view TCN (npairs)     & $8.0\%$ & $9.0\%$ & $35.9\%$ & $24.7\%$ & $35.5\%$ \\
multi-view TCN (triplet)    & $7.8\%$ & $10.0\%$ & $34.8\%$ & $22.7\%$ & $31.5\%$ \\
multi-view TCN (lifted)     & $7.8\%$ & $9.0\%$ & $35.4\%$ & $17.9\%$ & $27.7\%$ \\
\end{tabular}
\caption{
{\bf Detailed attributes classification errors,} for model selected by validation loss.}
\label{table:attributes_error}
\end{center}
\end{table}


\subsection{Learning Object Interaction Skills}
\label{sec:rl_experiments}
In this section, we use the TCN-based reward function described in \sect{reward_imitation} to learn robotic imitation behaviors from third-person demonstrations through reinforcement learning. We evaluate our approach on two tasks, plate transfer in a simulated dish rack environment (\fig{dishrack}, using the Bullet physics engine \cite{coumans2017}) and real robot pouring from human demonstrations (\fig{robot_pouring}). 

\subsubsection{Task Setup}
\begin{figure}[t]
\begin{center}
\centerline{
\includegraphics[width=.325\columnwidth]{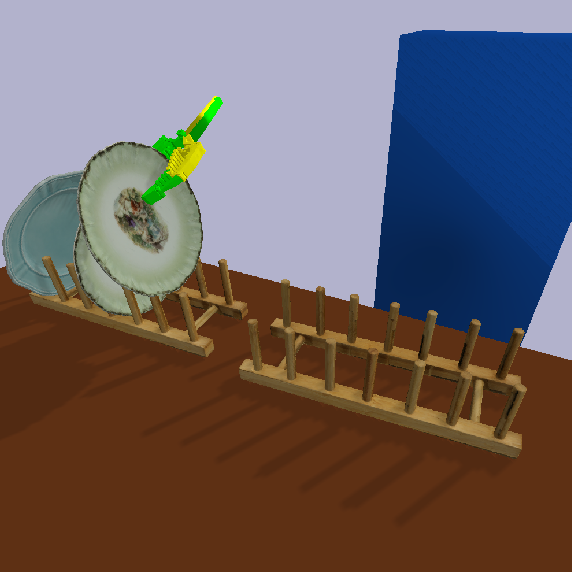}
\includegraphics[width=.325\columnwidth]{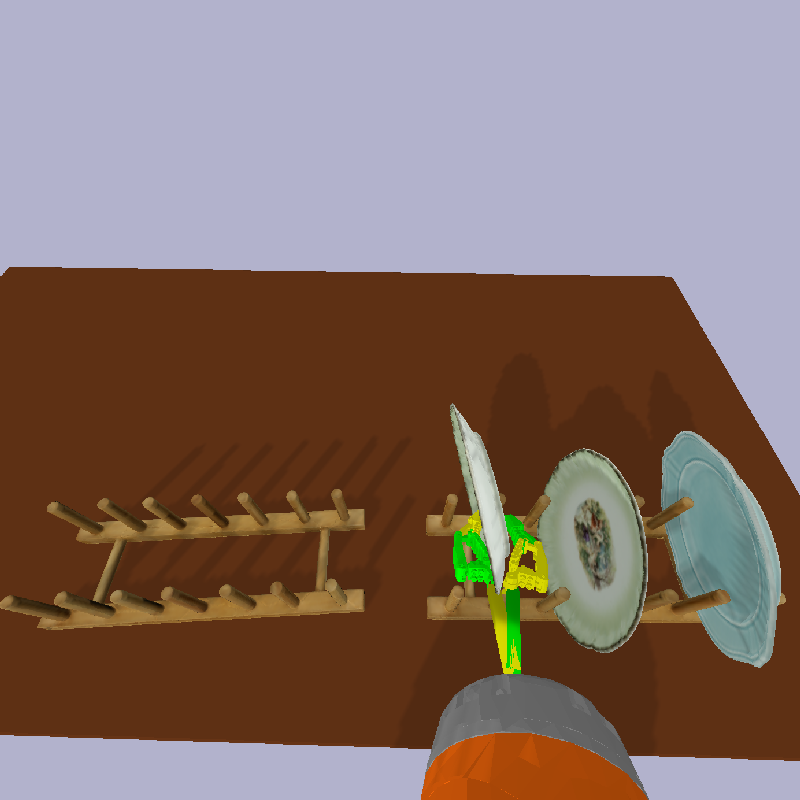}
\includegraphics[width=.35\columnwidth,trim={0 0 0.5cm 0},clip]{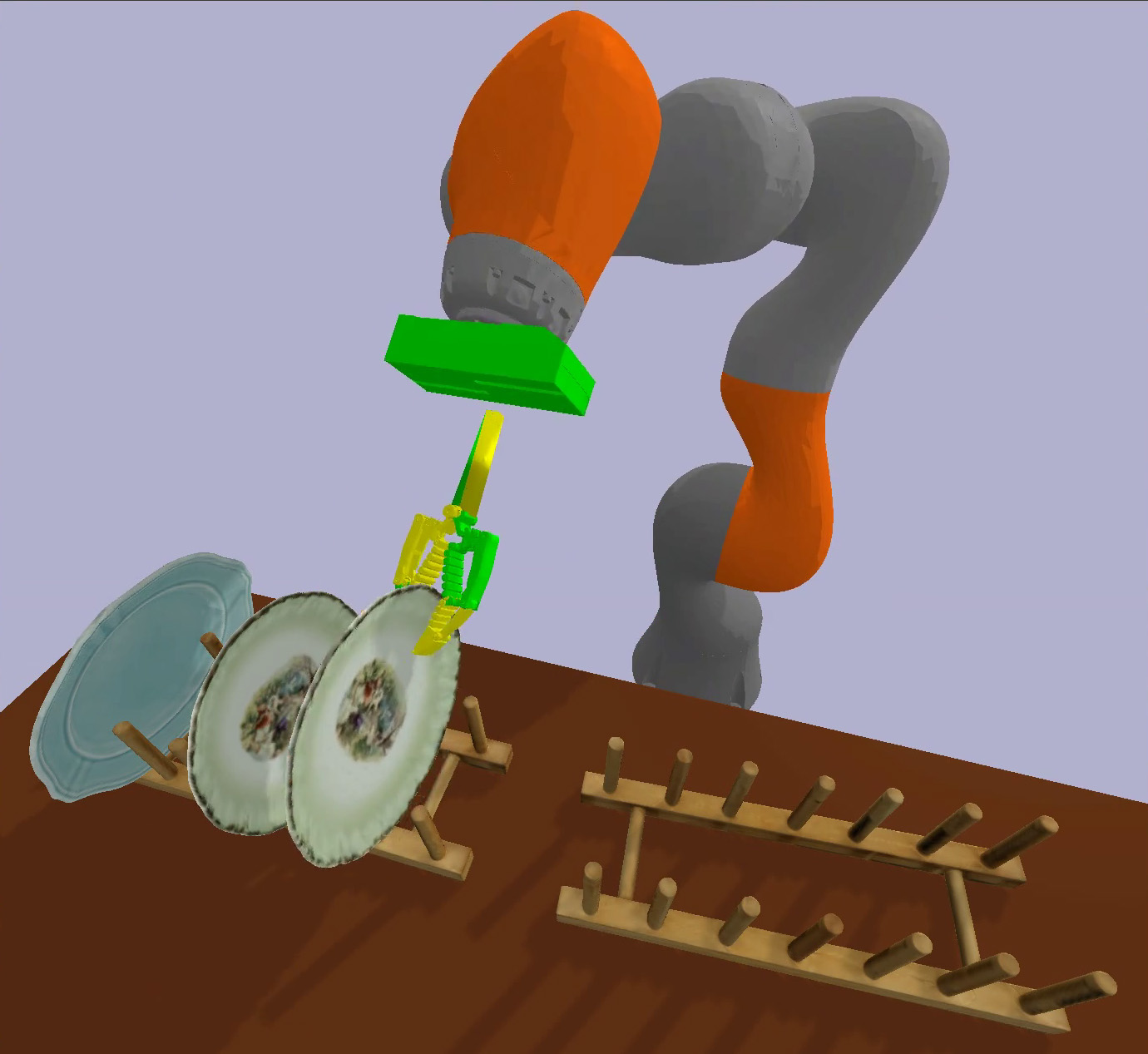}
}
\caption{{\bf Simulated dish rack task.} 
\textit{Left:} Third-person VR demonstration of the dish rack task. \textit{Middle:} View from the robot camera during training. \textit{Right:} Robot executing the dish rack task.
}
\vspace{-1.0em}
\label{fig:dishrack}
\end{center}
\end{figure}

The simulated dish rack environment consists of two dish racks placed on a table and filled with plates. The goal of the task is to move plates from one dish rack to another without dropping them. This requires a complex motion with multiple stages, such as reaching, grasping, picking up, carrying, and placing of the plate. We record the human demonstrations using a virtual reality (VR) system to manipulate a free-floating gripper and move the plates (\fig{dishrack} left). We record the videos of the VR demonstrations by placing first-view and third-person cameras in the simulated world. In addition to demonstrations, we also record a range of randomized motions  to increase the generalization ability of our TCN model. 
After recording the demonstrations, we place a simulated 7-DoF KUKA robotic arm inside the dish rack environment (\fig{dishrack} right) and attach a first-view camera to it. The robot camera images (\fig{dishrack} middle) are then used to compute the TCN reward function. The robot policy is initialized with random Gaussian noise.

\begin{figure}[t]
\begin{center}
\centerline{
\includegraphics[width=.2\columnwidth]{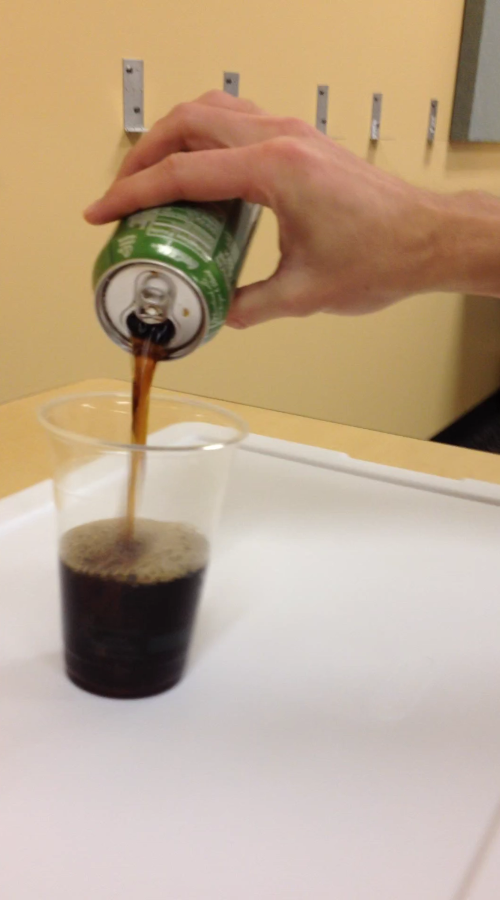}
\includegraphics[width=.4\columnwidth]{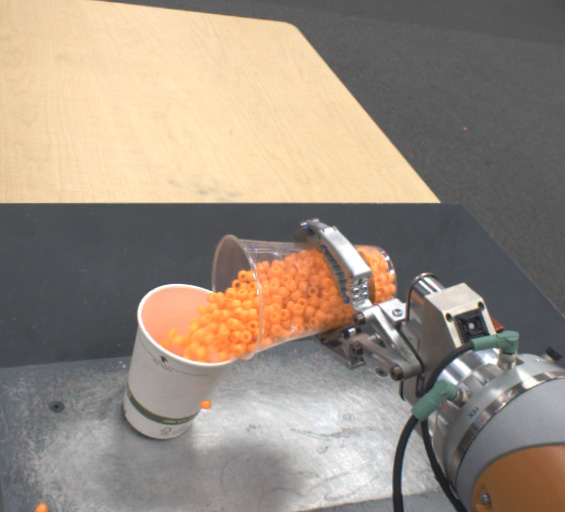}
\includegraphics[width=.4\columnwidth]{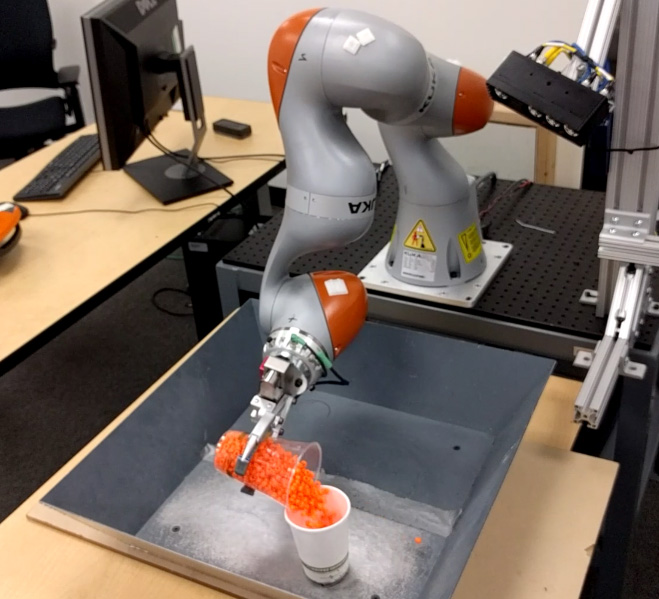}
}
\caption{{\bf Real robot pouring task.}
\textit{Left:} Third-person human demonstration of the pouring task. \textit{Middle:} View from the robot camera during training. \textit{Right:} Robot executing the pouring task.
}
\vspace{-1.0em}
\label{fig:robot_pouring}
\end{center}
\end{figure}

For the real robot pouring task, we first collect the multi-view data from multiple cameras to train the TCN model. The training set includes videos of humans performing pouring of liquids recorded on smartphone cameras and videos of robot performing pouring of granular beads recorded on two robot cameras.
We not only collect positive demonstrations of the task at hand, we also collect various interactions that do not actually involve pouring, such as moving cups around, tipping them over, spilling beads, etc, to cover the range of possible events the robot might need to understand.
The pouring experiment analyzes how TCNs can implicit build correspondences between human and robot manipulation of objects. The dataset that we used to train the TCN consisted of ${\sim}20$ minutes of humans performing pouring tasks, as well as ${\sim}20$ additional minutes of humans manipulating cups and bottles in ways other than pouring, such as moving the cups, tipping them over, etc. In order for the TCN to be able to represent both human and robot arms, and implicitly put them into correspondence, it must also be provided with data that allows it to understand the appearance of robot arms. To that end, we added data consisting of ${\sim} 20$ minutes of robot arms manipulating cups in pouring-like settings. Note that this data does not necessarily need to itself illustrate successful pouring tasks: the final demonstration that is tracked during reinforcement learning consists of a human successfully pouring a cup of fluid, while the robot performs the pouring task with orange beads. However, we found that providing some clips featuring robot arms was important for the TCN to acquire a representation that could correctly register the similarities between human and robot pouring. Using additional robot data is justified here because it would not be realistic to expect the robot to do well while having never seen its own arm. Over time however, the more tasks are learned the less needed this should become.
While the TCN is trained with approximately 1 hour of pouring-related multi-view sequences, the robot policy is only learned from a single liquid pouring video provided by a human (\fig{robot_pouring} left). With this video, we train a 7-DoF KUKA robot to perform the pouring of granular beads as depicted in \fig{robot_pouring} (right).
We compute TCN embeddings from the robot camera images (\fig{robot_pouring} middle) and initialize the robot policy using random Gaussian noise. We set the initial exploration higher on the wrist joint as it contributes the most to the pouring motion (for all compared algorithms).

\begin{figure}[t]
\begin{center}
\centerline{\includegraphics[width=1.0\columnwidth]{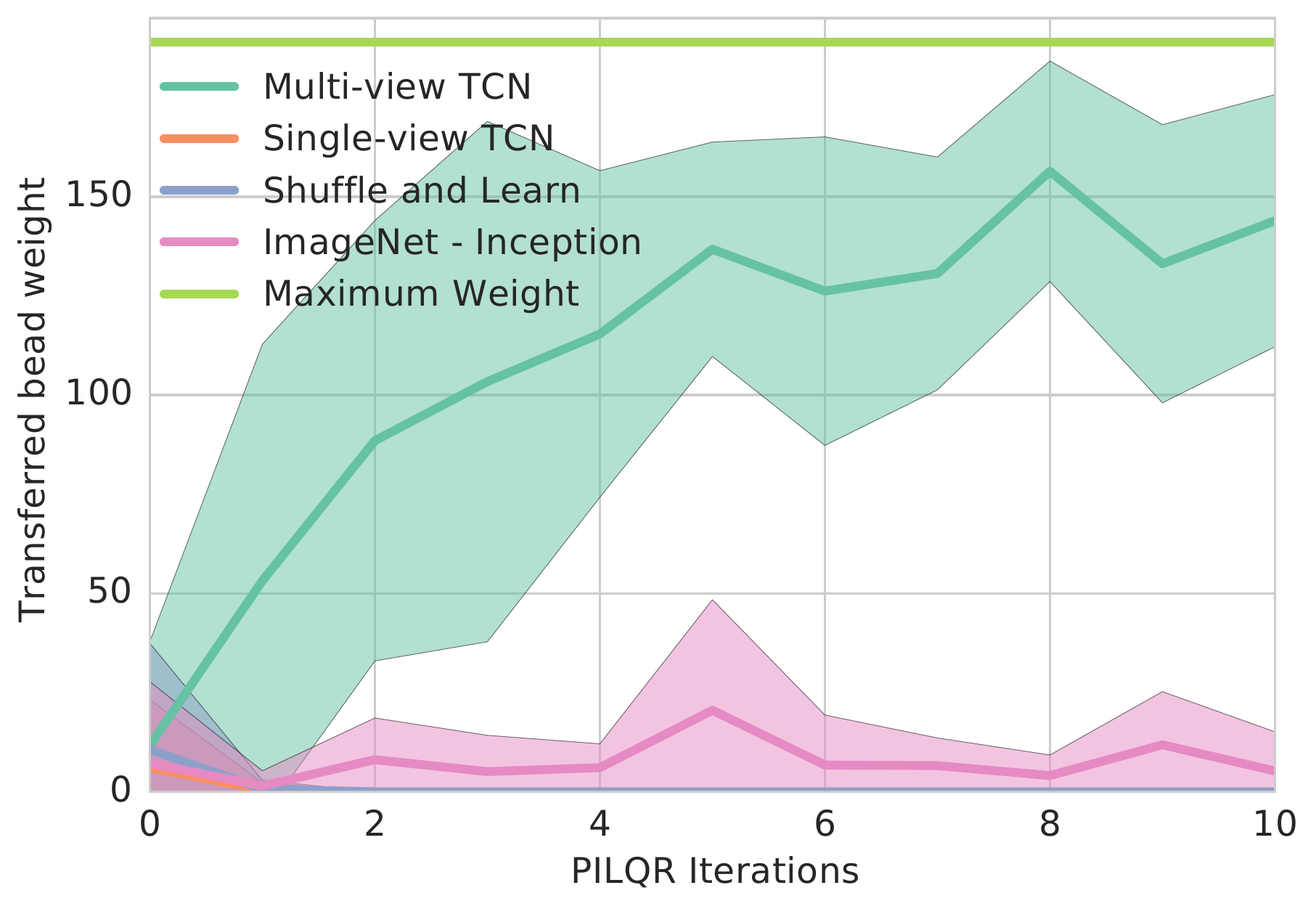}}
\vspace{-0.2cm}
\caption{
{\bf Learning progress of the pouring task,} using a single 3rd-person human demonstration, as shown in \fig{robot_pouring}. This graph reports the weight in grams measured from the target recipient after each pouring action (maximum weight is 189g) along with the standard deviation of all 10 rollouts per iteration. The robot manages to successfully learn the pouring task using the multi-view TCN model after only 10 iterations.
}
\vspace{-0.5cm}
\label{fig:bead_weights}
\end{center}
\end{figure} 

\subsubsection{Quantitative Evaluation}

\fig{bead_weights} shows the pouring task performance of using TCN models for reward computation compared to the same baselines evaluated in the previous section. After each roll-out, we measure the weight of the beads in the receiving container. We perform runs of 10 roll-outs per iteration. Results in \fig{bead_weights} are averaged over 4 runs per model (2 runs for 2 fixed random seeds). Already after the first several iterations of using the multi-view TCN model (mvTCN), the robot is able to successfully pour significant amount of the beads. After 10 iterations, the policy converges to a consistently successful pouring behavior. In contrast, the robot fails to accomplish the task with other models. Interestingly, we observe a low performance for single-view models (single-view TCN and shuffle \& learn) despite being trained on the exact same multi-view data as mvTCN. We observe the same pattern in \afig{bead_weights_demo2} when using a different human demonstration. This suggests taking advantage of multi-view correspondences is necessary in this task for correctly modeling object interaction from a 3rd-person perspective.
The results show that mvTCN does provide the robot with suitable guidance to understand the pouring task. In fact, since the PILQR~\cite{pilqr_icml17} method uses both model-based and model-free updates, the experiment shows that mvTCN not only provides good indicators when the pouring is successful, but also useful gradients when it isn't; while the other tested representations are insufficient to learn this task. This experiment illustrates how self-supervised representation learning and continuous rewards from visual demonstrations can alleviate the sample efficiency problem of reinforcement learning.

\subsubsection{Qualitative Evaluation}
As shown in our supplementary video, both dish rack and pouring policies converge to robust imitated behaviors. In the dish rack task, the robot is able to gradually learn all of the task components, including the arm motion and the opening and closing of the gripper. It first learns to reach for the plate, then grasp and pick it up and finally carry it over to another dish rack and place it there. The learning of this task requires only 10 iterations, with 10 roll-outs per iteration. This shows that the TCN reward function is dense and smooth enough to efficiently guide the robot to a complex imitation policy.

In the pouring task, the robot starts with Gaussian exploration by randomly rotating and moving the cup filled with beads. The robot first learns to move and rotate the cup towards the receiving container, missing the target cup and spilling large amount of the beads in the early iterations. After several more iterations, the robot learns to be more precise and eventually it is able to consistently pour most of the beads in the last iteration. This demonstrates that our method can efficiently learn tasks with non-linear dynamic object transitions, such as movement of the granular media and liquids, an otherwise difficult task to perform using conventional state estimation techniques.


\subsection{Self-Regression for Human Pose Imitation}
\label{sec:imitation_learning}

\begin{figure}[h]
\begin{center}
\centerline{\includegraphics[width=1\columnwidth]{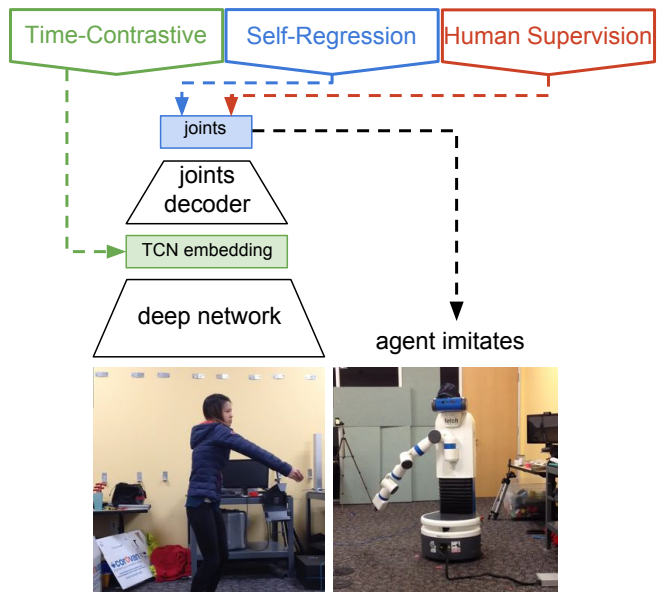}}
\caption{
{\bf TCN for self-supervised human pose imitation:} architecture, training and imitation. The embedding is trained unsupervised with the time-contrastive loss, while the joints decoder can be trained with self-supervision, human supervision or both. Output joints can be used directly by the robot planner to perform the imitation. Human pose is never explicitly represented.}
\vspace{-0.3cm}
\label{fig:self_regression_model}
\end{center}
\end{figure}

In the previous section, we showed that we can use the TCN to construct a reward function for learning object manipulation with reinforcement learning. In this section, we also study how the TCN can be used to directly map from humans to robots in real time, as depicted in \fig{self_regression_model}: in addition to understanding object interaction, we can use the TCN to build a pose-sensitive embedding either unsupervised, or with minimal supervision.
The multi-view TCN is particularly well suited for this task because, in addition to requiring viewpoint and robot/human invariance, the correspondence problem is ill-defined and difficult to supervise.
Apart from adding a joints decoder on top of the TCN embedding and training it with a self-regression signal, there is no fundamental difference in the method.
Throughout this section, we use the robot joint vectors corresponding to the human-to-robot imitation described in \fig{signals} as ground truth. Human images are fed into the imitation system, and the resulting joints vector are compared against the ground truth joints vector.
\vspace{-0.4em}

\begin{table}[h]
\small
\begin{center}
\begin{tabular}{l|c}
{\bf Supervision} & {\bf Robot joints distance error \%} \\
\hline
Random (possible) joints & $ 42.4 \pm 0.1$\\
Self & $38.8 \pm 0.1$\\
Human & $33.4 \pm 0.4$\\
Human + Self & $33.0 \pm 0.5$\\
TC + Self & $32.1 \pm 0.3$\\
TC + Human & $29.7 \pm 0.1$\\
TC + Human + Self & ${\bf 29.5} \pm 0.2$
\end{tabular}
\end{center}
\vspace{-.18cm}
\caption{{\bf Imitation error for different combinations of supervision signals.} The error reported is the joints distance between prediction and groundtruth. Note perfect imitation is not possible.}
\label{table:comparing_supervisions}
\vspace{.18cm}
\end{table}

By comparing different combinations of supervision signals, we show in \tabl{comparing_supervisions} that training with all signals performs best. We observe that adding the time-contrastive signal always significantly improves performance. In general, we conclude that relatively large amounts of cheap weakly-supervised data and small amounts of expensive human supervised data is an effective balance for our problem. Interestingly, we find that the self-supervised model (TC+self) outperforms the human-supervised one.
It should however be noted that the quantitative evaluation is not as informative here: since the task is highly subjective and different human subjects imitate the robot differently, matching the joint angles on held-out data is exceedingly difficult. We invite the reader to watch the accompanying video for examples of imitation, and observe that there is a close connection between the human and robot motion, including for subtle elements of the pose such as crouching: when the human crouches down, the robot lowers the torso via the prismatic joint in the spine. In the video, we observe a complex human-robot mapping is discovered entirely without human supervision. This invites to reflect on the need for intermediate human pose detectors when correspondence is ill-defined as in this case. In \afig{tsne_zoom}, we visualize the TCN embedding for pose imitation and show that pose across humans and robots is consistent within clusters, while being invariant to viewpoint and backgrounds. More analysis is available in \appen{pose_imitation_analysis}.

\section{Conclusion}

In this paper, we introduced a self-supervised representation learning method (TCN) based on multi-view video. The representation is learned by anchoring a temporally contrastive signal against co-occuring frames from other viewpoints, resulting in a representation that disambiguates temporal changes (e.g., salient events) while providing invariance to viewpoint and other nuisance variables. We show that this representation can be used to provide a reward function within a reinforcement learning system for robotic object manipulation, and to provide mappings between human and robot poses to enable pose imitation directly from raw video. In both cases, the TCN enables robotic imitation from raw videos of humans performing various tasks, accounting for the domain shift between human and robot bodies. Although the training process requires a dataset of multi-viewpoint videos, once the TCN is trained, only a single raw video demonstration is used for imitation. Limitations and future work are discussed in \appen{future}.

\subsubsection*{Acknowledgments}
{\small
We thank Mohi Khansari, Yunfei Bai, Erwin Coumans, Jonathan Tompson, James Davidson and Vincent Vanhoucke for helpful feedback and Ashwin Kakarla for helping labeling evaluation data. We thank everyone who provided imitations for this project: Phing Lee, Alexander Toshev, Anna Goldie, Deanna Chen, Deirdre Quillen, Dieterich Lawson, Eric Langlois, Ethan Holly, Irwan Bello, Jasmine Collins, Jeff Dean, Julian Ibarz, Ken Oslund, Laura Downs, Leslie Phillips, Luke Metz, Mike Schuster, Ryan Dahl, Sam Schoenholz and Yifei Feng.
}

\vspace{-.2cm}
\bibliographystyle{unsrtnat} 
\renewcommand{\bibfont}{\small}
\bibliography{icra}

\ifbool{isArxiv}{}{
\newpage
\clearpage
}
\appendix

\subsection{Future Work}
\label{appendix.future}
One of the limitations of our approach is that the representation requires multi-viewpoint video for training, which is not as widely available (e.g. from the Internet) as standard video. We do analyze a single-viewpoint variant of our approach, and find that it also achieves improvement over baseline ImageNet-trained features, but that the multi-viewpoint TCN achieves substantially better results. However, as robots become more ubiquitous, recording multi-viewpoint video autonomously, for example by using stereo cameras, seems like a promising direction. Our method can also be viewed as a specific instance of a more general class of multi-modal embedding methods, where temporally co-occuring events in multiple sensory modalities are embedded into the same space. Viewed in this light, the exploration of broader modalities, such as audio and tactile sensing, would be an exciting avenue for future work.
Another limitation in our experiments is that we train a separate TCN for each task (pouring, pose imitation, etc.). While the TCN for a given task, such as pouring, is trained on videos that include clips of failed pouring, moving cups, and so on, the embedding is only used to learn a single task, namely pouring. In principle, the embeddings learned by the TCN should be task agnostic, though considerably larger datasets may be required in this case. An interesting avenue for future work would be to study how multiple tasks could be embedded in the same space, essentially creating a universal representation for imitation of object interaction behaviors.
While in this paper we explored learning representations using time as a supervision signal, in the future, models should learn simultaneously from a collection of diverse yet complementary signals.

\subsection{Reinforcement Learning Details}
\label{appendix.rl}
Let $p(\mathbf{u}_t | \mathbf{x}_t)$ be a robot policy, which defines a probability distribution over 
actions $\mathbf{u}_t$ conditioned on the system state $\mathbf{x}_t$ at each time step $t$ of a task execution. We employ policy search to optimize the policy parameters $\theta$. Let $\tau = (\mathbf{x}_1, \mathbf{u}_1, \dots,  \mathbf{x}_T, \mathbf{u}_T)$ be a trajectory of states and actions. Given a cost function\footnote{We use a cost function instead of the reward function as it is more common in the trajectory optimization theory.} $c(\mathbf{x}_t, \mathbf{u}_t)$, we define the trajectory cost as $c(\tau)=\sum_{t=1}^T c(\mathbf{x}_t, \mathbf{u}_t)$. The policy is optimized with respect to the expected cost of the policy
\[
J(\theta) = \mathbb{E}_{p}\left[c(\tau)\right] = \int c(\tau) p (\tau) d\tau\,,
\]
where $p(\tau)$ is the policy trajectory distribution given the system dynamics $p\left(\mathbf{x}_{t+1} | \mathbf{x}_{t}, \mathbf{u}_{t}\right)$
\[
p (\tau) = p(\mathbf{x}_1) \prod_{t=1}^{T} p\left(\mathbf{x}_{t+1} | \mathbf{x}_{t}, \mathbf{u}_{t}\right) p(\mathbf{u}_t | \mathbf{x}_t)\,.
\]

One policy class that allows us to employ very efficient reinforcement learning methods is the time-varying linear-Gaussian (TVLG) controller $p(\mathbf{u}_t | \mathbf{x}_t) = \mathcal{N}(\mathbf{K}_{t} \mathbf{x}_t + \mathbf{k}_{t}, \mathbf{\Sigma}_{t})$. In this work, we apply a reinforcement learning method PILQR\acite{pilqr_icml17} to learn these TVLG controllers on a real robot, which combines model-based and model-free policy updates for an efficient learning of tasks with complex system dynamics. 

Let  $S(\mathbf{x}_{t},\mathbf{u}_{t}) = c(\mathbf{x}_{t}, \mathbf{u}_{t}) + \sum^T_{j=t+1} c(\mathbf{x}_{j}, \mathbf{u}_{j})$ be the cost-to-go of a trajectory starting in state $\mathbf{x}_{t}$ by performing action $\mathbf{u}_{t}$ and following the policy $p(\mathbf{u}_t | \mathbf{x}_t)$ afterwards. In each iteration $i$, PILQR performs a KL-constrained optimization of $S(\mathbf{x}_{t},\mathbf{u}_{t})$:
\begin{align*}
\!\min_{p^{(i)}}~\mathbb{E}_{p^{(i)}}[S(\mathbf{x}_{t},\mathbf{u}_{t})]~
s.t.~\mathbb{E}_{p^{(i-1)}}\!\!\left[D_\text{KL} \left(p^{(i)}\|\, p^{(i-1)}\right)\right]\leq \epsilon,
\end{align*}
where limiting the KL-divergence between the new policy $p^{(i)}$ and the old policy $p^{(i-1)}$ leads to a better convergence behavior. The optimization is divided into two steps. In the first step, we perform a fast model-based update using an algorithm LQR-FLM~\acite{LevineA14}, which is based on the iterative linear-quadratic regulator (iLQR)~\acite{synthesis} and approximates $S(\mathbf{x}_{t},\mathbf{u}_{t})$ with a linear-quadratic cost $\hat{S}(\mathbf{x}_{t},\mathbf{u}_{t})$. In the second step, the residual cost-to-go $\tilde{S}(\mathbf{x}_{t},\mathbf{u}_{t}) = S(\mathbf{x}_{t},\mathbf{u}_{t}) - \hat{S}(\mathbf{x}_{t},\mathbf{u}_{t})$ is optimized using a model-free method PI$^2$~\acite{TheodorouBS10,chebotar-icra2017} to produce an unbiased policy update.

\subsection{Objects Interaction Analysis}
\label{appendix.objects_interaction_analysis}

Here, we visualize the embeddings from the ImageNet-Inception and multi-view TCN models with t-SNE using a coloring by groundtruth labels. Each color is a unique combination of 5 attribute values defined earlier, i.e. if each color is well separated the model can identify uniquely all possible combinations of our 5 attributes. Indeed we observe in \afig{tsne_classcombinations} some amount of color separation for the TCN but not for the baseline.

\begin{table}[h]
\small
\begin{center}
\begin{tabular}{l|c|c|c}
{\bf Method} & {\bf alignment} & {\bf classif.} & {\bf training} \\
& {\bf error} & {\bf error} & {\bf iteration}\\
\hline
Random & $28.1\% $ & $54.2\%$ & -\\
Inception-ImageNet & $ 29.8\% $ & $ 51.9\%$ & -\\
single-view TCN (triplet) & $ 26.6\% $ & $ 23.6\% $& 738k\\
shuffle \& learn~\citep{DBLP:journals/corr/MisraZH16} & $ 20.9\% $ & $ 23.2\%$ & 743k\\
multi-view TCN (lifted) & $ 19.1\% $ & $21.4\%$ & 927k\\
multi-view TCN (triplet) & $17.5\% $ & $ 20.3\%$ & 47k\\
multi-view TCN (npairs) & $ 17.5\% $ & $19.3\%$ & 224k\\
\end{tabular}
\caption{\textbf{Pouring alignment and classification errors: } these models are selected using the classification score on a small labeled validation set, then ran on the full test set. We observe that multi-view TCN outperforms other models with 15x shorter training time. The classification error considers 5 classes related to pouring: "hand contact with recipient", "within pouring distance", "container angle", "liquid is flowing" and "recipient fullness".}
\label{table:appendix.order_metric_selected_from_classification}
\end{center}
\end{table}

\begin{figure}[htb]
\begin{center}
\centerline{\includegraphics[width=\columnwidth]{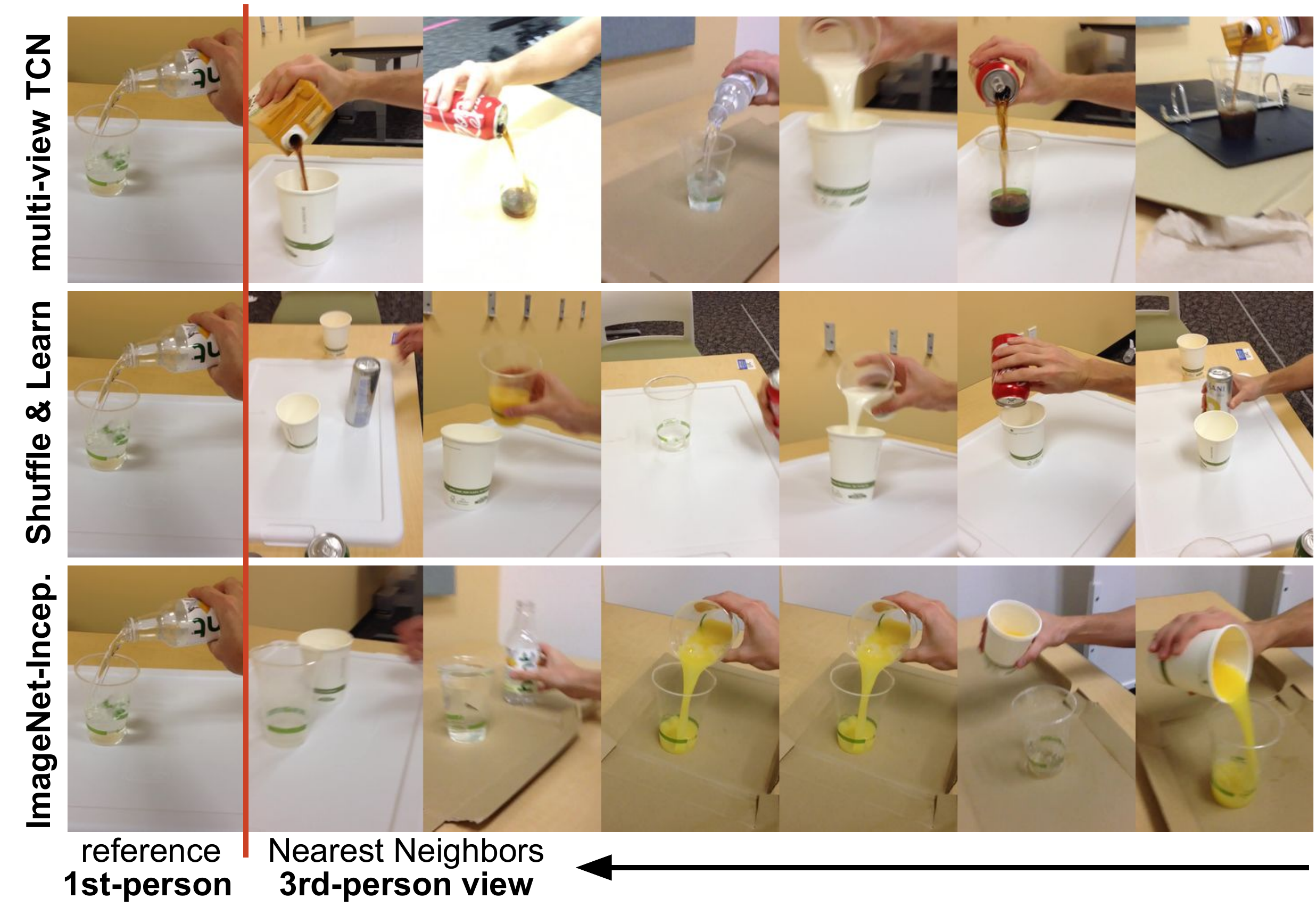}}
\caption{{\bf Label-free pouring imitation}: nearest neighbors (right) for each reference image (left) for different models (multi-view TCN, shuffle \& learn and ImageNet-Inception). These pouring test images show that {\bf the TCN model can distinguish different hand poses and amounts of poured liquid simply from unsupervised observation} while being invariant to viewpoint, background, objects and subjects, motion-blur and scale.}
\label{fig:appendix.knn_pouring}
\end{center}
\end{figure} 

\begin{figure}[htb]
\begin{center}
\centerline{\includegraphics[width=\columnwidth]{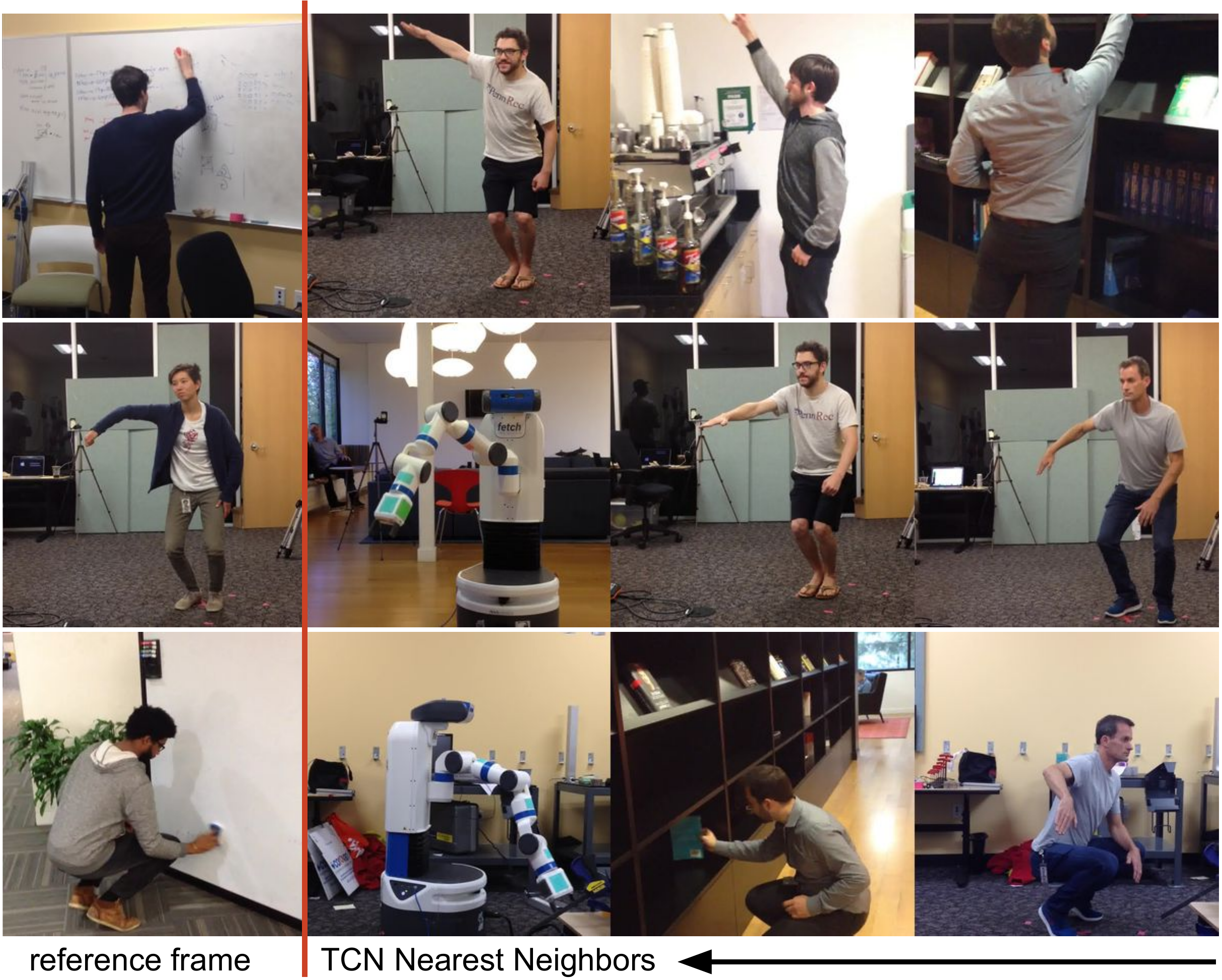}}
\caption{{\bf Label-free pose imitation}: nearest neighbors (right) for each reference frame (left) for each row. Although only trained with self-supervision (no human labels), the multi-view TCN can understand correspondences between humans and robots for poses such as crouching, reaching up and others while being invariant to viewpoint, background, subjects and scale.}
\label{fig:appendix.knn_pose}
\end{center}
\end{figure} 

\begin{figure}[h]
\begin{center}
\centerline{\includegraphics[width=.8\columnwidth]{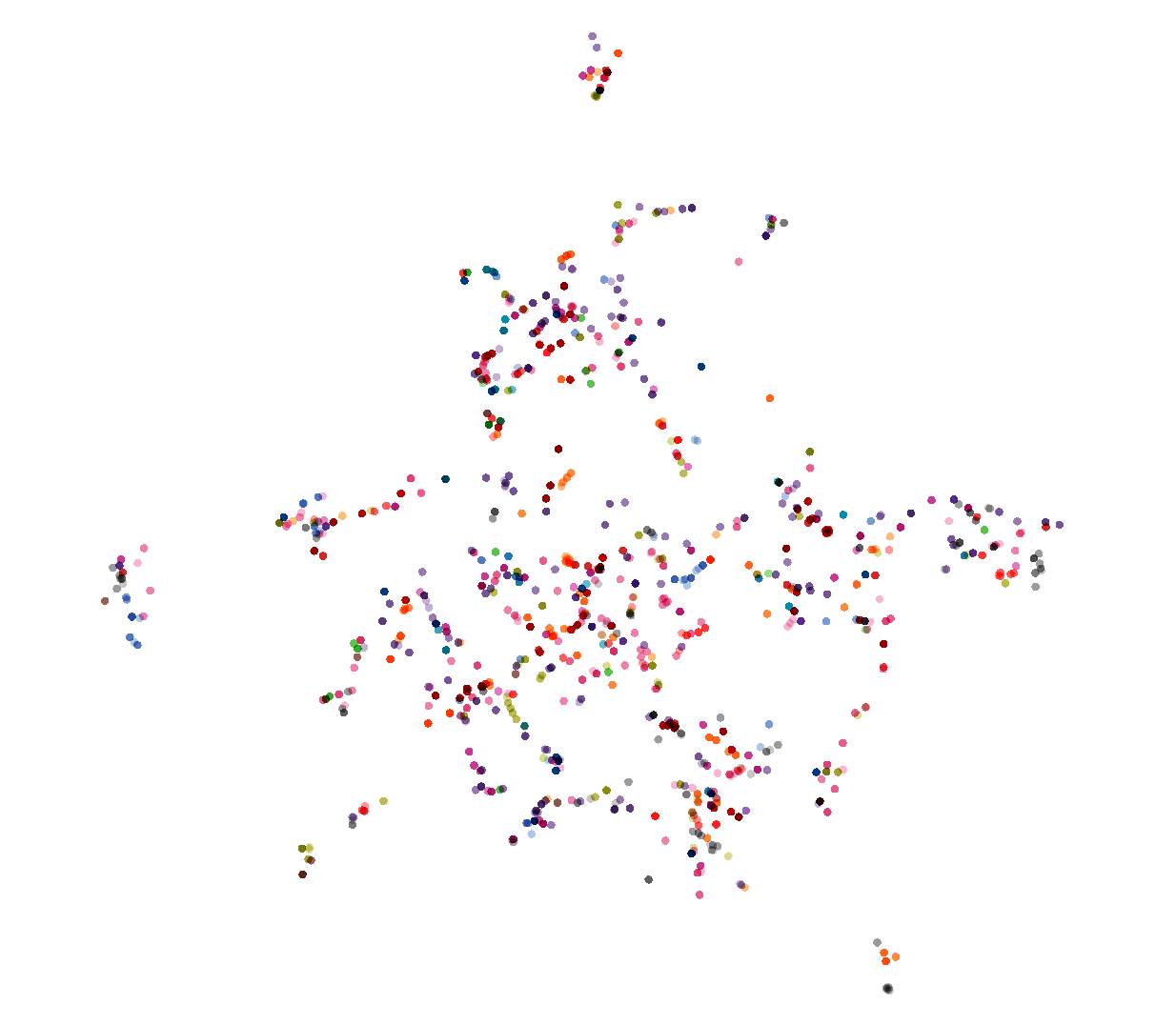}}
\centerline{\includegraphics[width=.8\columnwidth]{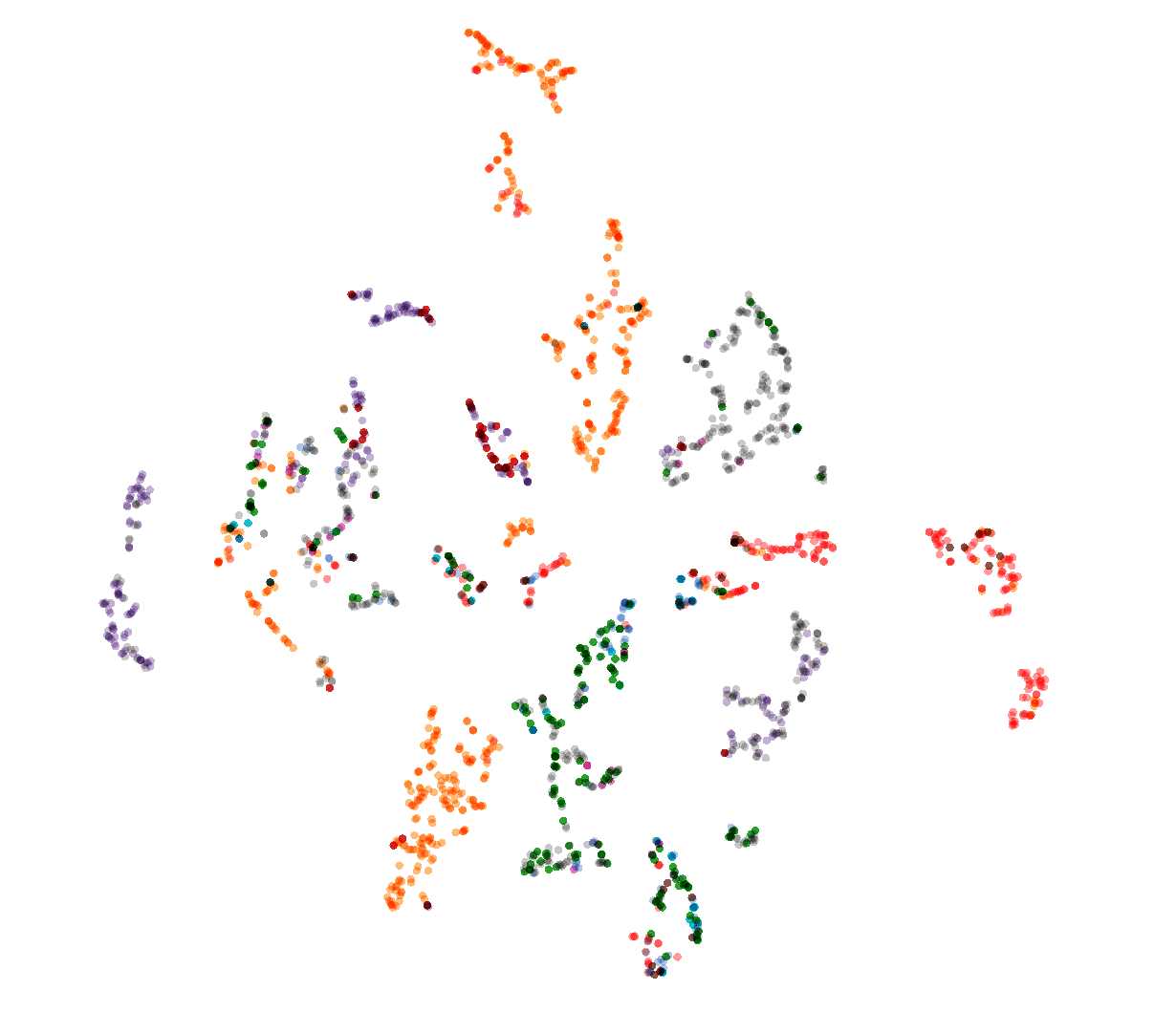}}
\caption{
{\bf t-SNE colored by attribute combinations:} TCN (bottom) does a better job than ImageNet-Inception (top) at separating combinations of attributes.}
\label{fig:appendix.tsne_classcombinations}
\end{center}
\end{figure} 

\begin{figure}[t]
\begin{center}
\centerline{\includegraphics[width=1.0\columnwidth]{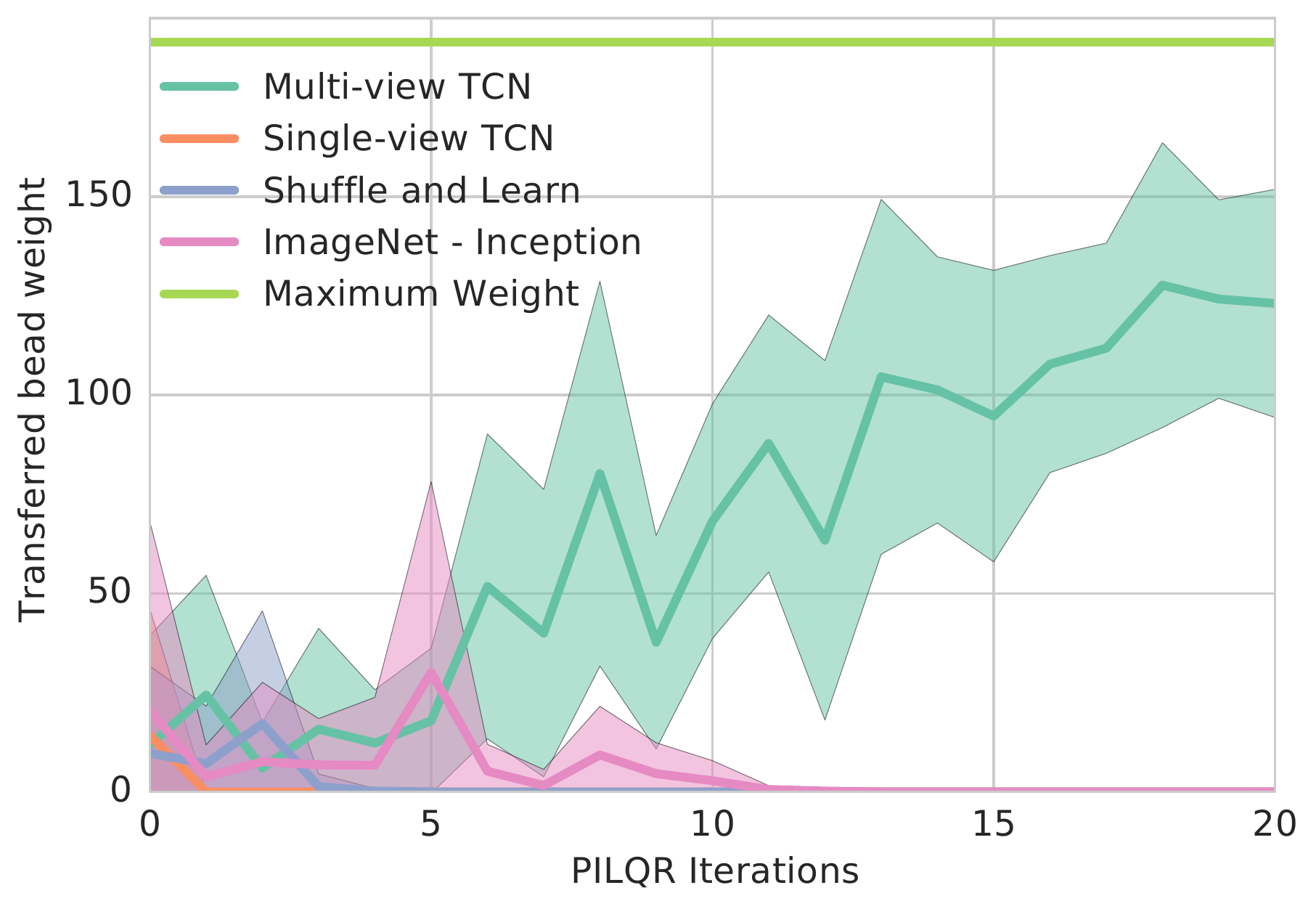}}
\vspace{-0.2cm}
\caption{
{\bf Learning progress of the pouring task,} using a single 3rd-person human demonstration that is different that the one shown in \fig{robot_pouring}. This graph reports the weight in grams measured from the target recipient after each pouring action (maximum weight is 189g) along with the standard deviation of all 10 rollouts per iteration. The robot manages to successfully learn the pouring task using the multi-view TCN model after 20 iterations.
}
\vspace{-0.5cm}
\label{fig:appendix.bead_weights_demo2}
\end{center}
\end{figure}

\newpage
\subsection{Pose Imitation Analysis}
\label{appendix.pose_imitation_analysis}

\subsubsection{Pose Imitation Data}
\label{appendix.pose_data}

The human training data consists of sequences distinguished by human subject and clothing pair. Each sequence is approximately 4 minutes. For the label-free TC supervision we collected approximately 30 human pairs (about 2 hours) where humans imitate a robot but the joint labels are not recorded, along with 50 robot sequences with random motion (about 3 hours, trivial to collect). For human supervision, we collected 10 human/clothing pairs (about 40 minutes, very expensive collection) while also recording the joints labels. Each recorded sequence is captured by 3 smartphone cameras fixed on tripods at specific angles ($0^{\circ}$, $60^{\circ}$ and $120^{\circ}$) and distance. The validation and testing sets each consist of 6 human/clothing pairs not seen during training (about 24 minutes, very expensive collection).

The distance error is normalized by the full value range of each joint, resulting in a percentage error. Note that the Human Supervision signal is quite noisy, since the imitation task is subjective and different human subjects interpret the mapping differently. In fact, a perfect imitation is not possible due to physiological differences between human bodies and the Fetch robot. Therefore, the best comparison metric available to us is to see whether the joint angles predicted from a held-out human observation match the actual joint angles that the human was attempting to imitate.

\subsubsection{Models}
\label{appendix.models_imitation}

We train our model using the 3 different signals as described in \fig{signals}.
The model consists of a TCN as described in \sect{models}, to which we add a joint decoder network (2 fully-connected layers above TC embedding: $\rightarrow 128 \rightarrow 8$, producing 8 joint values). We train the joints decoder with L2 regression using the self-supervision or human supervision signals. The model can be trained with different combinations of signals; we study the effects of each combination in section \afig{graph_supervisions}. The datasets used are approximately 2 hours of random human motions, 3 hours of random robot motions and 40 minutes of human supervision, as detailed in \appen{pose_data}. At test time, the resulting joints vector can then directly be fed to the robot stack to update its joints (using the native Fetch planner) as depicted in \fig{self_regression_model}. This results in an end-to-end imitation from pixels to joints without any explicit representation of human pose.

\subsubsection{Supervision Analysis}
\label{sec:comparing_supervision_amounts}

\begin{figure}[h]
\begin{center}
\centerline{\includegraphics[width=\columnwidth]{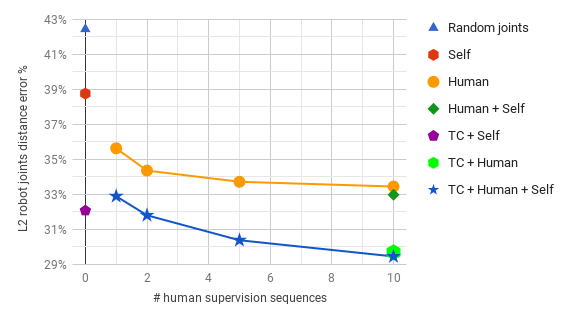}}
\vspace{-.1cm}
\caption{
{\bf Comparing types and amounts of supervision:} Self-supervised imitation ("TC+self") outperforms human-supervised imitation ("Human").}
\label{fig:appendix.graph_supervisions}
\end{center}
\vspace{-.5cm}
\end{figure} 

As shown in \fig{self_regression_model}, our imitation system can be trained with different combinations of signals. Here we study how our self-supervised imitation system compares to the other possible combinations of training signals. The performance of each combination is reported in \tabl{comparing_supervisions} using the maximum amounts of data available (10 sequences for Human supervision and 30 sequences for TC supervision), while \afig{graph_supervisions} varies the amount of human supervision. Models such as the "TC + Self" or "Self" do not make use of any human supervision, hence only appear as single points on the vertical axis. Models that do not include TC supervision are simply trained as end-to-end regression problems. For example the “Self” model is trained end-to-end to predict internal joints from third person observations of the robot, and then that model is applied directly to the human imitation task.
For reference, we compute a random baseline which samples joints values within physically possible ranges. In general, we observe that more human supervision decreases the L2 robot joints error. It is interesting to note that while not given any labels, the self-supervised model ("TC + Self") still significantly outperforms the fully-supervised model ("Human"). The combination of all supervision signals performs the best. Overall, we observe that adding the TC supervision to any other signal significantly decreases the imitation error. In \afig{graph_unsup}, we vary the amount of TC supervision provided and find the imitation error keeps decreasing as we increase the amount of data. Based on these results, we can make the argument that relatively large amounts of cheap weakly-supervised data and small amounts of expensive human supervised data is an effective balance for our problem. A non-extensive analysis of viewpoint and scale invariance in \sect{imitation_invariance} seems to indicate that the model remains relatively competitive when presented with viewpoints and scales not seen during training. 

\subsubsection{Analysis by Joint}

In \afig{graph_joints}, we examine the error of each joint individually for 4 models. Interestingly, we find that for all joints excepts for "shoulder pan", the unsupervised "TC+Self" models performs almost as well as the human-supervised "TC+Human+Self". The unsupervised model does not seem to correctly model the shoulder pan and performs worse than Random. Hence most of the benefits of human supervision found in \afig{graph_supervisions} come from correcting the shoulder pan prediction.

\subsubsection{Qualitative Results}

We offer multiple qualitative evaluations: k-Nearest Neighbors (kNN) in \afig{knn_pose}, imitation strips in \afig{strips} and a t-SNE visualization in \afig{tsne_zoom}. Video strips do not fully convey the quality of imitation, we strongly encourage readers to watch the videos accompanying this paper.
{\bf kNN:} In \afig{knn_pose}, we show the nearest neighbors of the reference frames for the self-supervised model "TC+Self" (no human supervision). Although never trained across humans, it learned to associate poses such as crouching or reaching up between humans never seen during training and with entirely new backgrounds, while exhibiting viewpoint, scale and translation invariance.
{\bf Imitation strips:} In \afig{strips}, we present an example of how the self-supervised model has learned to imitate the height level of humans by itself (easier to see in supplementary videos) using the "torso" joint (see \afig{graph_joints}). This stark example of the complex mapping between human and robot joints illustrates the need for learned mappings, here we learned a non-linear mapping from many human joints to a single "torso" robot joint without any human supervision.
{\bf t-SNE:} We qualitatively evaluate the arrangement of our learned embedding using t-SNE representations with perplexity of $30$ and learning rate of $10$. In \afig{tsne_zoom}, we show that the agent-colored embedding exhibits local coherence with respect to pose while being invariant to agent and viewpoint.
More kNN examples, imitation strips and t-SNE visualizations from different models are available in \sect{imitation_examples}.

\begin{figure}[h]
\begin{center}
\centerline{\includegraphics[width=1\columnwidth]{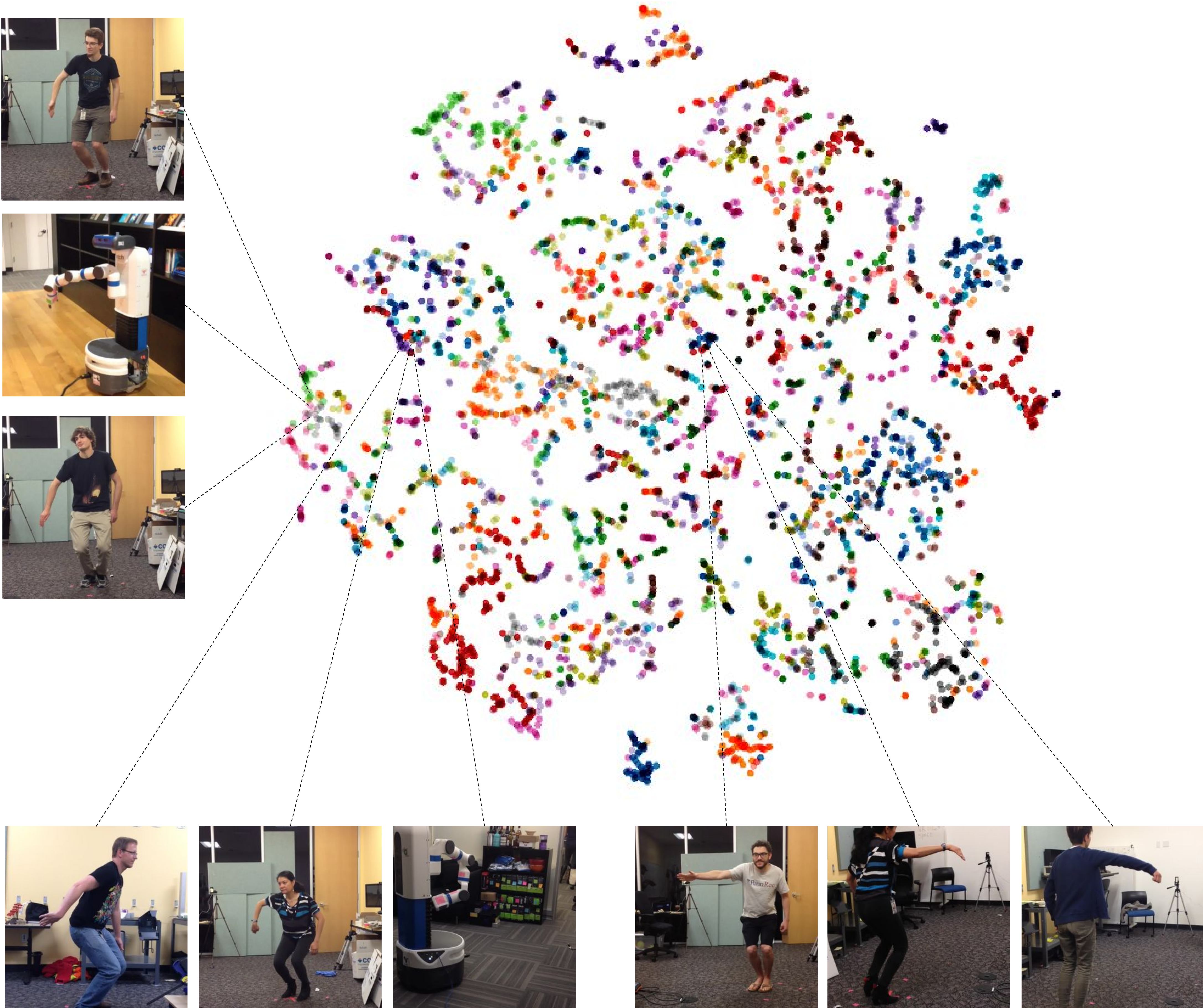}}
\caption{{\bf t-SNE embedding colored by agent for model "TC+Self".}
We show that images are locally coherent with respect to pose while being invariant to agent or viewpoint.}
\vspace{-1.0em}
\label{fig:appendix.tsne_zoom}
\end{center}
\end{figure} 

\begin{figure}[h]
\begin{center}
\centerline{\includegraphics[width=\columnwidth]{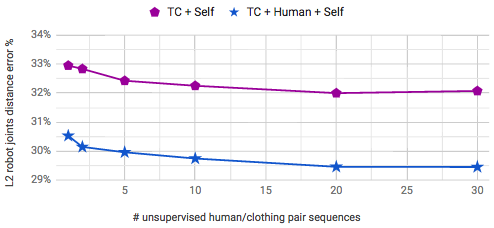}}
\caption{
{\bf Varying the amount of unsupervised data:} increasing the number of unsupervised sequences decreases the imitation error for both models.}
\label{fig:appendix.graph_unsup}
\end{center}
\end{figure} 

\begin{figure}[h]
\begin{center}
\centerline{\includegraphics[width=\columnwidth]{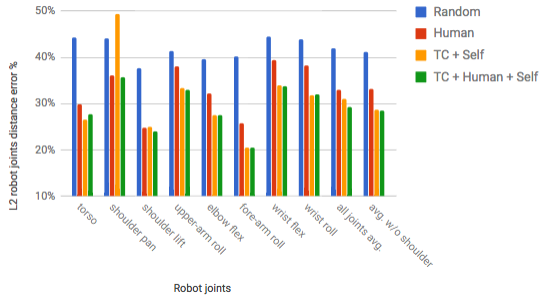}}
\caption{
{\bf L2 robot error break-down by robot joints.} From left to right, we report errors for the 8 joints of the Fetch robot, followed by the joints average, followed by the joints average excluding the "shoulder pan" join.}
\label{fig:appendix.graph_joints}
\end{center}
\end{figure}

\begin{figure}[h]
\begin{center}
\centerline{\includegraphics[width=.8\columnwidth]{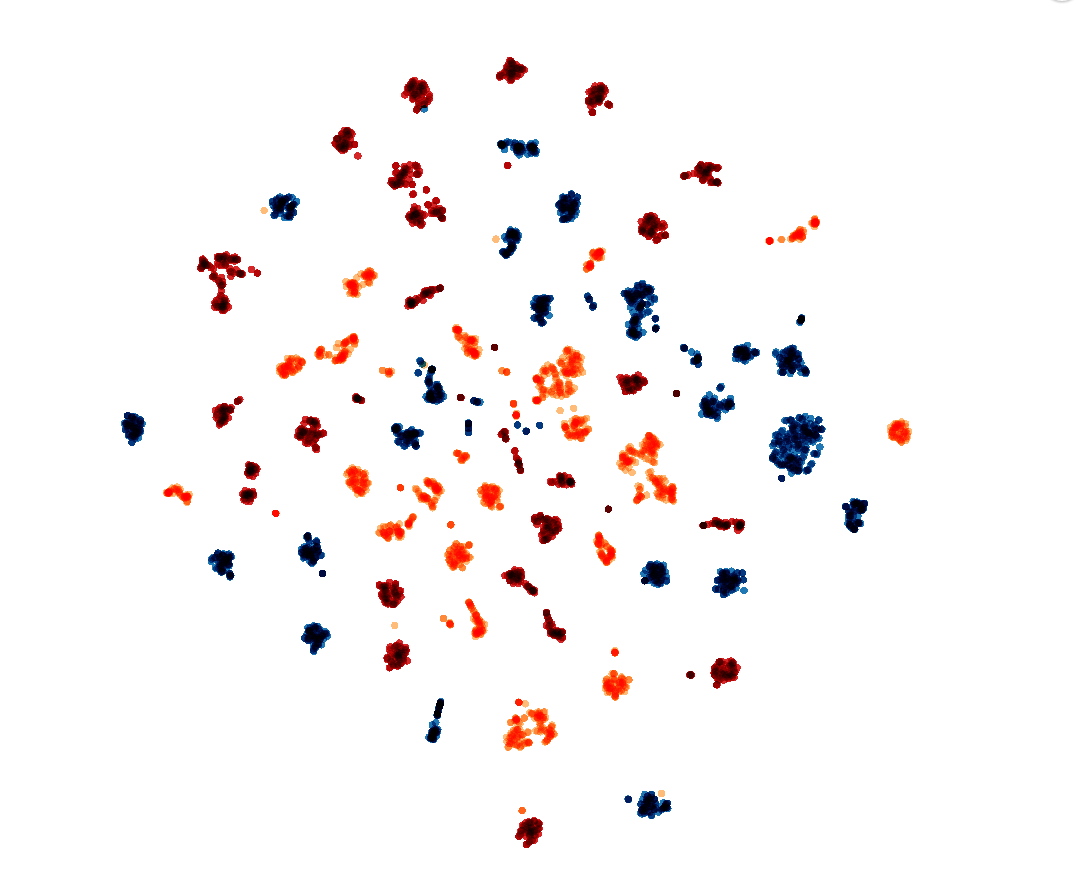}}
\centerline{\includegraphics[width=.8\columnwidth]{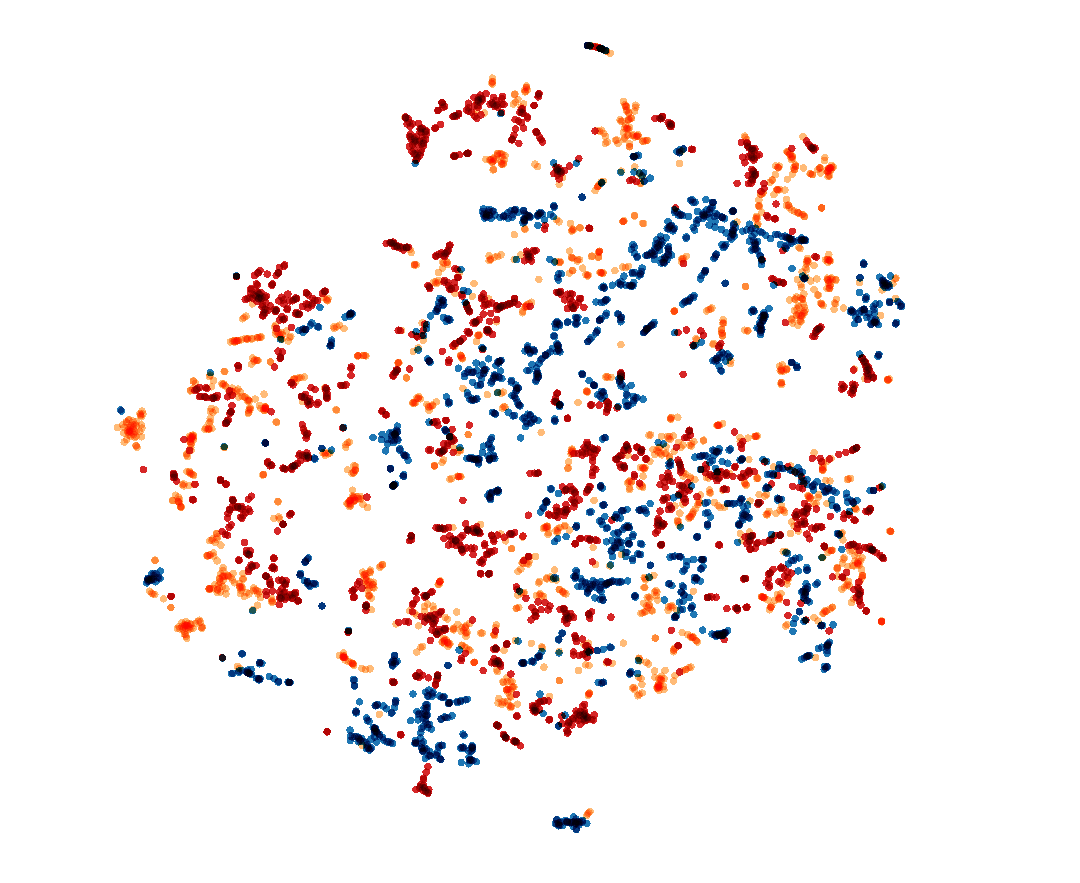}}
\caption{{\bf t-SNE embedding before (top) and after (bottom) training, colored by view.}
Before training, we observe concentrated clusters of the same color, indicating that the manifold is organized in a highly view-specific way, while after training each color is spread over the entire manifold.
}
\label{fig:appendix.bridge_tsne_view}
\end{center}
\end{figure} 

\begin{figure}[h]
\begin{center}
\centerline{\includegraphics[width=.8\columnwidth]{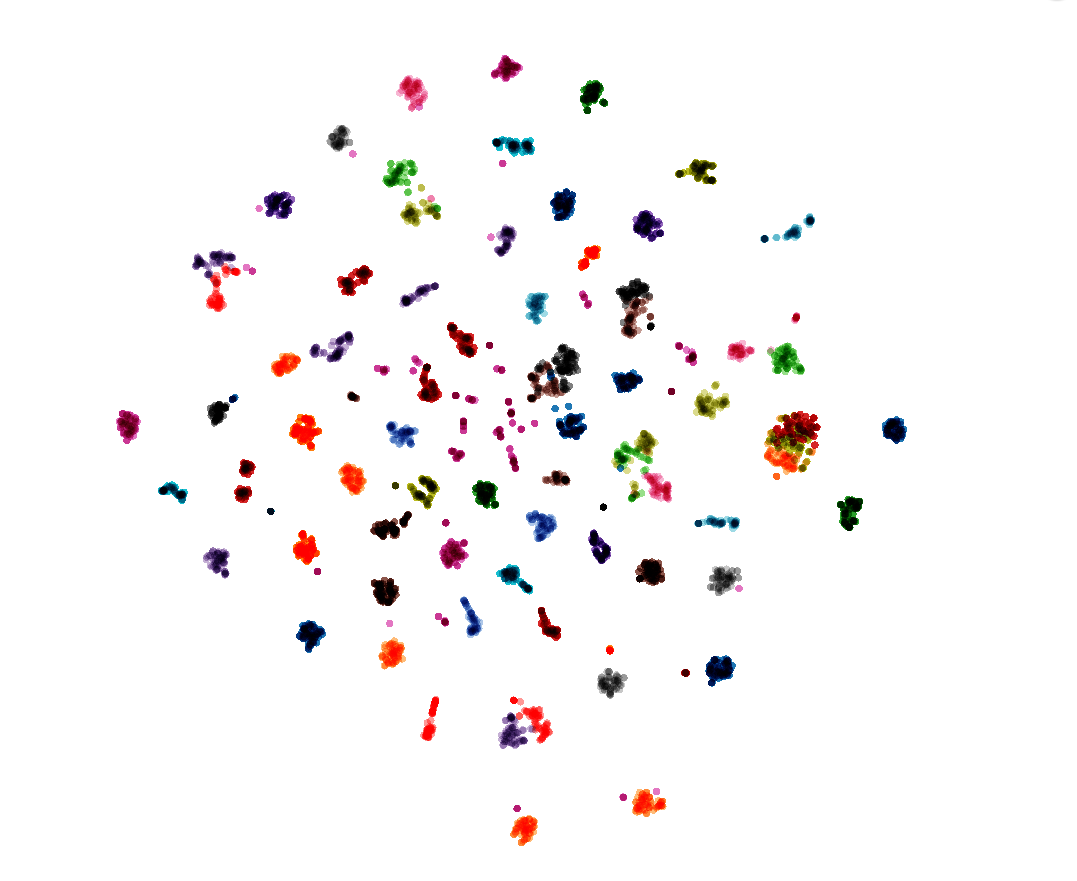}}
\centerline{\includegraphics[width=.8\columnwidth]{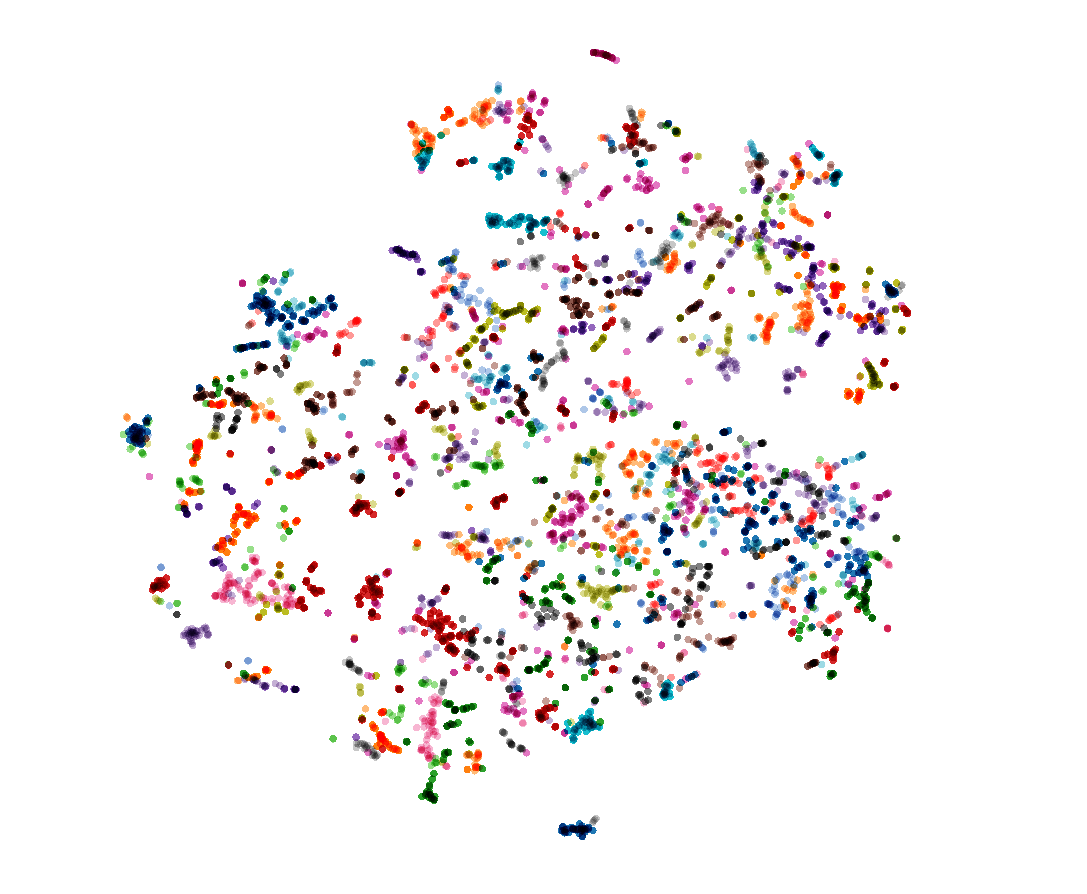}}
\caption{{\bf t-SNE embedding before (top) and after (bottom) training, colored by agent.}
Before training, we observe concentrated clusters of the same color, indicating that the manifold is organized in a highly agent-specific way, while after training each color is spread over the entire manifold.}
\label{fig:appendix.bridge_tsne_agent}
\end{center}
\end{figure} 

\subsection{Imitation Invariance Analysis}
\label{sec:imitation_invariance}

\begin{figure}[h]
\begin{center}
\centerline{\includegraphics[width=\columnwidth]{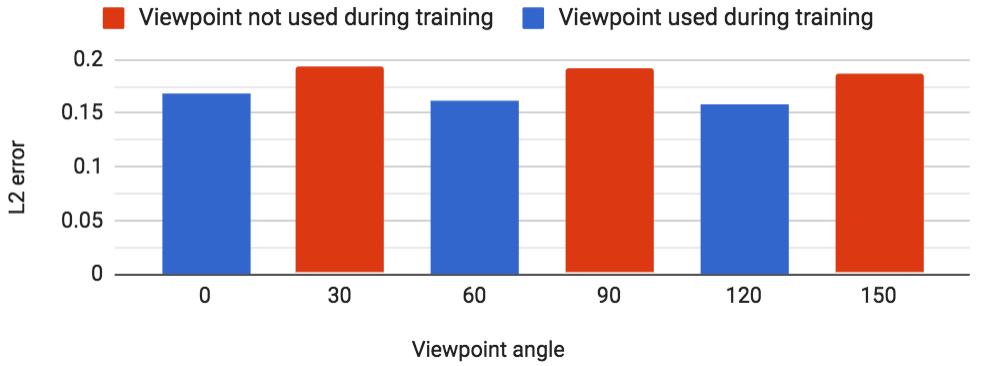}}
\caption{
{\bf Testing TC+Human+Self model for orientation invariance:} while the error increases for viewpoints not seen during training  ($30^{\circ}$, $90^{\circ}$ and $150^{\circ}$), it remains competitive.
}
\label{fig:graph_angles}
\end{center}
\end{figure}

\begin{figure}[h]
\begin{center}
\centerline{\includegraphics[width=\columnwidth]{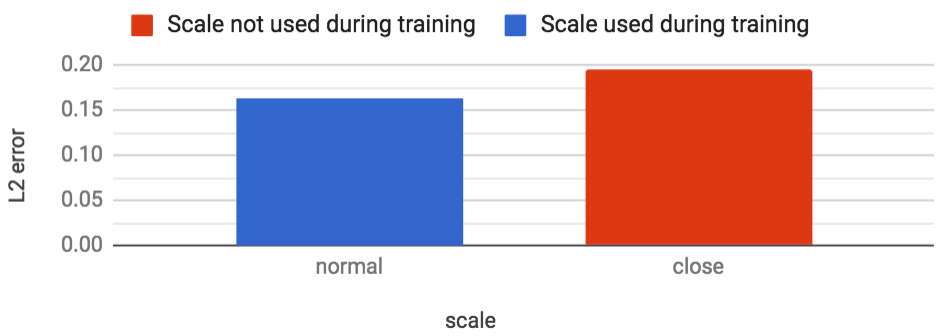}}
\caption{
{\bf Testing for scale invariance:} while the error increases when decreasing the distance of the camera to the subject (about half way compared to training), it remains competitive and lower than the human-supervised baseline.}
\label{fig:graph_distance}
\end{center}
\end{figure}

In this section, we explore how much invariance is captured by the model. In \fig{graph_angles}, we test the L2 imitation error from new viewpoints ($30^{\circ}$, $90^{\circ}$ and $150^{\circ}$) different from training viewpoints ($0^{\circ}$, $60^{\circ}$ and $120^{\circ}$). We find that the error increases but the model does not break down and keeps a lower accuracy than the Human model in \afig{graph_supervisions}. We also evaluate in \fig{graph_distance} the accuracy while bringing the camera closer than during training (about half way) and similarly find that while the error increases, it remains competitive and lower than the human supervision baseline. From these experiments, we conclude that the model is somewhat robust to viewpoint changes (distance and orientation) even though it was trained with only 3 fixed viewpoints.

\subsection{Imitation Examples}
\label{sec:imitation_examples}

\begin{figure*}[htb!]
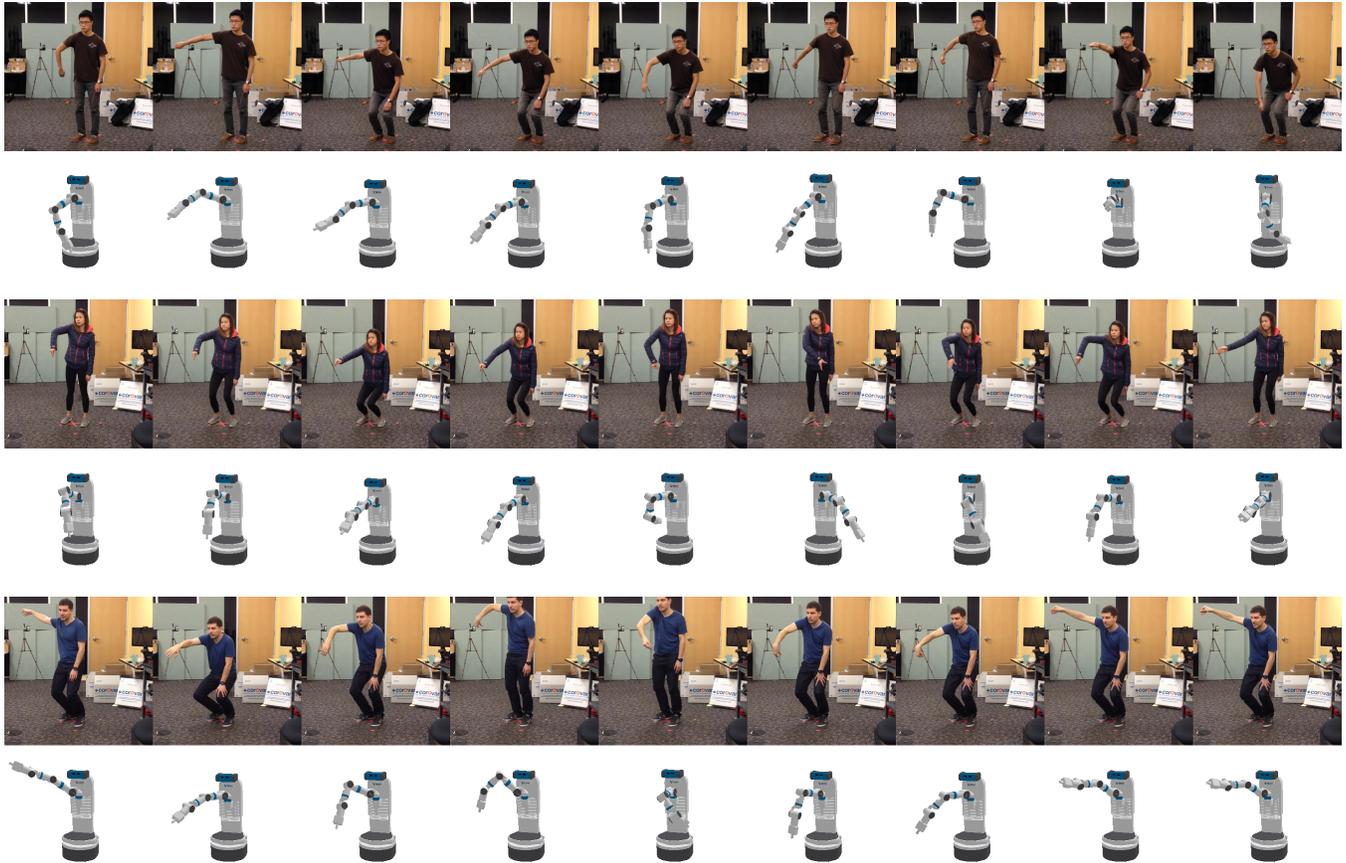

\begin{center}
\centerline{\includegraphics[width=1\linewidth]{\faceblur{strips/strip_eric}}}
\centerline{\includegraphics[width=1\linewidth]{\faceblur{strips/strip_phing}}}
\centerline{\includegraphics[width=1\linewidth]{\faceblur{strips/strip_pierre}}}
\caption{
{\bf Self-supervised imitation examples.} Although not trained using any human supervision (model "TC+Self"), the TCN is able to approximately imitate human subjects unseen during training. Note from the rows (1,2) that the TCN discovered the mapping between the robot's torso joint (up/down) and the complex set of human joints commanding crouching. In rows (3,4), we change the capture conditions compared to training (see rows 1 and 2) by using a free-form camera motion, a close-up scale and introduction some motion-blur and observe that imitation is still reasonable.}
\label{fig:appendix.strips}
\end{center}
\end{figure*} 


{\bf t-SNE:} We qualitatively evaluate the arrangement of our learned embedding using t-SNE representations with perplexity of $30$ and learning rate of $10$. In this section we show the embedding before and after training, and colorize points by agent in \fig{appendix.bridge_tsne_agent} and by view in \fig{appendix.bridge_tsne_view}. The representations show that the embedding initially clusters views and agents together, while after training points from a same agent or view spread over the entire manifold, indicating view and agent invariance.


\end{document}